\documentclass[times, twoside]{zHenriquesLab-StyleBioRxiv}
\usepackage{blindtext}
\usepackage{multirow}
\usepackage{epsfig}
\usepackage{multirow}
\usepackage{graphicx}
\usepackage{amsmath}
\usepackage{amssymb}
\usepackage{tabu}
\usepackage{booktabs}
\usepackage{lipsum}
\usepackage{longtable}
\usepackage{lineno}
\newcommand{\beginsupplement}{%
        \setcounter{table}{0}
        \renewcommand{\thetable}{S\arabic{table}}%
        \setcounter{figure}{0}
        \renewcommand{\thefigure}{S\arabic{figure}}%
     }

\leadauthor{Tison} 

\begin{document}

\title{Automated and Interpretable Patient ECG Profiles for Disease Detection, Tracking, and Discovery}
\shorttitle{Automated and Interpretable ECG Patient Profiles}


\author[1,*]{Geoffrey H. Tison}
\author[2,3,*]{Jeffrey Zhang}
\author[1]{Francesca N. Delling}
\author[1,2,4,5,6,7,8\Letter]{Rahul C. Deo}

\affil[1]{Division of Cardiology, Department of Medicine, University of California, San Francisco, CA, USA.}
\affil[2]{Cardiovascular Research Institute, University of California, San Francisco, CA, USA.}
\affil[3]{Department of Electrical Engineering and Computer Science, University of California at Berkeley, CA, USA.}
\affil[4]{Institute for Human Genetics, University of California, San Francisco, CA, USA.}
\affil[5]{California Institute for Quantitative Biosciences, San Francisco, USA.}
\affil[6]{Institute for Computational Health Sciences, University of California, San Francisco, USA.}
\affil[7]{Center for Digital Health Innovation, University of California, San Francisco, USA.}
\affil[8]{Present address:  One Brave Idea and Division of Cardiology, Department of Medicine, Brigham and Women's Hospital.}
\affil[*]{contributed equally}
\maketitle
\bfabstractfalse

\begin{abstract}
\textbf{Background:}  The electrocardiogram or ECG has been in use for over 100 years and remains the most widely performed diagnostic test for characterization of cardiac structure and electrical activity. Remarkably, current approaches to automated ECG interpretation originate from heuristics devised over 40 years ago. 
textbf{Objective:}  We hypothesized that parallel advances in computing power, innovations in machine learning algorithms, and availability of large-scale digitized ECG data would enable extending the utility of the ECG beyond its current limitations, while at the same time preserving interpretability, an attribute which remains critical to medical decision-making.
\textbf{Methods:} We identified 36,186 ECGs from the UCSF database that were 1) in normal sinus rhythm and 2) would enable training of specific models for estimation of cardiac structure or function or detection of disease. We derived a novel model for ECG segmentation using convolutional neural networks (CNN) and Hidden Markov Models (HMM) and evaluated its output by comparing electrical interval estimates to 141,864 measurements produced during the clinical workflow. We built a 725-element patient-level ECG profile using downsampled ECG segmentation data and trained machine learning models to estimate left ventricular mass, left atrial volume, mitral annulus e' and to detect and track four diseases: pulmonary arterial hypertension (PAH), hypertrophic cardiomyopathy (HCM), cardiac amyloid (CA), and mitral valve prolapse (MVP).
\textbf{Results:} CNN-HMM derived ECG segmentation agreed with clinical estimates, with median absolute deviations (MAD) as a fraction of observed value of 0.6\% for heart rate, 3\% for PR interval, 4\% for QT interval, and 6\% for QRS duration. Patient-level ECG profiles enabled quantitative estimates of left ventricular mass (MAD vs. echocardiogram of 16\%) and mitral annulus e' velocity (MAD of 19\%) with good discrimination in binary classification models of left ventricular hypertrophy and diastolic dysfunction [Area Under the Receiver Operating Characteristic Curve (AUROC) of 0.87 and 0.84, respectively]. Models for disease detection ranged from AUROC of 0.94 for PAH, 0.91 for HCM, 0.86 for CA, and 0.77 for MVP. Top-ranked variables for all models included known ECG characteristics along with novel predictors of these traits/diseases. Furthermore, temporal variation in model-derived disease scores coincided with visual evolution of ECG morphologies for these features.
\textbf{Conclusion:} Modern AI methods can extend the 12-lead ECG to quantitative and diagnostic applications well beyond its current uses. Moreover, careful selection of machine learning algorithms achieves the goal of automation and accuracy without compromising the transparency that is so fundamental to clinical care and scientific discovery.

\end{abstract}

\begin{keywords}
electrocardiogram | machine learning | diagnosis | tracking
\end{keywords}

\begin{corrauthor}
rahul.c.deo\at gmail.com
\end{corrauthor}

Short Title: Automated, interpretable ECG analysis

\section*{Introduction}
The electrocardiogram (ECG) is the most commonly performed cardiovascular diagnostic procedure, with more than 100 million ECGs obtained annually in the United States \cite{Drazen:1988ud}, including use in 21\% of annual health examinations \cite{Bhatia:2017ji} and 17\% of emergency department visits \cite{Pitts:2008vw}. The ECG tracing is a direct reflection of underlying cardiac physiology, since its morphologic and temporal features are produced from cardiac electrical and structural variations. The paradigm of ECG interpretation has remained largely unchanged for decades: both physicians and computer algorithms apply specific rules — initially established by empiric, manual analysis and codified by clinical guidelines — to interrogate the ECG tracing for evidence of underlying disease \cite{BLACKBURN:1960un}.

Although modern ECG interpretation emphasizes binary classification (e.g. left ventricular hypertrophy or not), given the multiple physiologic and structural correlates of ECG signals, ECGs could be trained to estimate continuous parameters (i.e. a regression model), including structural and functional attributes of the heart. Moreover, for both classification and regression tasks, modern algorithms should also identify which components of the ECG signal drive their performance. Such an approach would provide the foundation not only to enable the discovery of new associations between the ECG signal and disease pathology, but also provide the transparency needed to reassure physicians and patients about the basis and validity of any automated diagnosis or parameter estimate. 

Since their introduction over 40 years ago \cite{PhD:2016et}, computerized algorithms have assisted physicians in ECG interpretation and are largely based upon the same expert-designed rules used by physicians. These rules have themselves been largely unchanged for decades, and derive from empiric, manual analysis of ECGs from various disease cohorts \cite{Kligfield:2007ir}. Analysis performed in this way can only evaluate simple heuristics on a small subset of the total information contained in an ECG, and has led to the familiar menu of criteria by which ECGs are evaluated, such as an R-wave >12mm in lead aVL suggesting left ventricular hypertrophy, according to the modified Cornell criteria \cite{MD:1985ck}. This traditional approach to ECG analysis does not readily account for high-level interactions between ECG signals from multiple leads, or small visually imperceptible yet informative changes which may exist in the signal, particularly in early disease stages. 

Novel techniques to analyze digital ECG data at large-scale would be foundational toward the goals of both improving algorithmic ECG interpretation and identifying novel ECG correlates of cardiac disease beyond existing criteria, all within a low-cost structure. A physician’s ability to track disease could also be substantially augmented by the ability to monitor and integrate subtle changes in serial ECGs. Presently, there does not exist an automated, scalable, algorithmic method to perform detailed longitudinal tracking and comparison of ECGs.

Machine learning algorithms have recently demonstrated revolutionary performance in the fields of computer vision \cite{Krizhevsky:2012wl} and speech recognition \cite{vandenOord:2016uo}, and more recently in medical applications \cite{Gulshan:2016iu,Esteva:2017ct}, but many of these innovative models suffer from being largely uninterpretable \cite{Lipton:2016tw}. In high-stakes fields such as medicine, this limits the ability to understand successes or troubleshoot failures, potentially dampening physician adoption of an unfamiliar technology.

We aimed to develop and test an algorithmic framework that facilitates scalable analysis of ECG data, while preserving interpretable parallels to cardiac physiology. This approach aspires to expand the flexibility and scalability of algorithmic ECG analysis, laying the crucial foundation to perform a wide range of novel ECG-based tasks including improving accuracy, estimating quantitative cardiac traits, performing longitudinal tracking of serial ECGs, and monitoring disease progression and risk.

\section*{Materials and Methods}

The source code for this project, including model weights, is available at https://bitbucket.org/rahuldeo/ecgai/.

\subsection{Human Subjects Research} University of California, San Francisco (UCSF) institutional review board approval was obtained for this study.

\subsection{Overview: Automated and interpretable ECG profiling for disease detection, tracking and discovery} We sought to develop an automated, scalable, and interpretable method to characterize 1) cardiac structure and 2) diastolic function; and 3) detect and track disease using patient-specific ECG profiles. Figure \ref{fig:figure1} demonstrates the analysis pipeline, data inputs and number of ECGs that were used in each step of algorithm development and validation. We termed the entire approach as ecgAI - referring to "artificial intelligence".

\subsection{ECG Data}
Standard 12-lead ECG data from 2010-2017 was obtained in XML format from the University of California, San Francisco (UCSF) clinical MUSE ECG database (MUSE Version 9.0 SP4, GE Healthcare, Wauwatosa, WI). Based on accompanying clinical and echocardiographic (echo) information (described below) we selected 36,186 ECGs, from which raw ECG voltage data was extracted for each of the 12 individual leads recorded over 10 seconds; 60\% of data was sampled at a frequency of 500Hz, and 40\% was sampled at 250Hz. As part of routine clinical care, each clinical ECG undergoes initial analysis by the GE software (MAC 5500 HD, Version 10, Revision F; Marquette 12SL; GE Healthcare, Wauwatosa, WI), and the interpretation is subsequently changed or confirmed by a UCSF cardiologist. We extracted standard ECG GE MUSE measurements, as well as final cardiologist-confirmed ECG diagnostic interpretations. Data from the UCSF electronic health record was obtained for relevant patients, including medical diagnoses, medications, specialty clinic referrals, and echo measurements.

\subsection{Selection of studies for model development}
We selected a subset of ECGs to train models for estimation of cardiac structure and function and detection of disease. To facilitate model development, we restricted the analyses to those ECGs for which the GE/UCSF rhythm interpretation was normal sinus rhythm.

For cardiac structure models, we searched the UCSF echo database for all instances of patients with echos and ECGs collected within 30 days of one another and who had recorded measurements either of left ventricular mass or left atrial volume. We found 10082 (Table \ref{tab:tableS1}) and 8289 (Table \ref{tab:tableS2}) studies, respectively, that met these criteria.  For cardiac diastolic function, we performed a similar search and found 4205 instances of patients with an ECG and a recorded mitral annulus medial e' value on echo within 30 days of each other (Table \ref{tab:tableS3}). There were fewer instances of lateral e' values recorded within our database and we thus focused our efforts on the medial e' metric.

We selected four diseases for which to perform a clinical demonstration of automated detection and tracking of disease using patient ECG profiles: pulmonary arterial hypertension (PAH), hypertrophic cardiomyopathy (HCM), cardiac amyloidosis (CA) and mitral valve prolapse (MVP). We previously identified the PAH, HCM and CA patients as part of a parallel study on developing a computer vision pipeline for automated echo interpretation \cite{Zhang:wd}. Briefly, on chart review HCM patients met guideline-based criteria \cite{Gersh:2011goa}; CA patients had both echo evidence of hypertrophy and confirmation of amyloidosis by biopsy or imaging; and PAH patients had an echo-indication of PAH and were on one of four PAH specific medications. MVP patients were identified by querying the UCSF echo database for patients with single or bileaflet MVP. Echo studies were subsequently over-read by a second board-certified cardiologist to confirm the diagnosis. We selected all ECGs corresponding to these patients that were available in XML format. To build classification models, we also matched each ECG to up to five ECGs matched by age (in 10 years bins), sex, year of study and race (the patient demographic information for ECGs in our archive has been organized in a python dictionary to facilitate the control selection process). Patient and study characteristics are described in Tables \ref{tab:tableS4}, \ref{tab:tableS5}, \ref{tab:tableS6}, and \ref{tab:tableS7}.

\subsection{ecgAI: a machine learning based approach to ECG segmentation}

To develop novel models to extend the utility of ECGs, we needed an efficient way to derive patient-specific ECG profiles, vectors of uniform length that capture the variation in ECG voltage over different leads. This first required a method to segment ECGs into their different components.

Historically, ECG segmentation has involved the application of a discrete or continuous wavelet transform to the raw ECG signal to help identify peaks \cite{Martinez:2004ge}. Typically, the QRS complex is located first, and then heuristics are applied (e.g. march backwards no more than a certain number of milliseconds until the signal diminishes below a certain threshold) to determine the onset of the QRS and the onset and termination of the P and T waves. Additional heuristics can be introduced to deal with abnormal heart rhythms such as atrial fibrillation or premature beats as well as difficult to detect P-waves and abnormally shaped QRS complexes.

Although effective, such heuristic-based approaches tend to be challenging to develop, as one must enumerate the exceptions to this initial approach and devise new rules to accommodate them. In this work, we explored the development of an alternative, technically novel approach to train an ECG segmentation model, capitalizing on recent advances in machine learning in the fields of computer vision and signal processing.

\subsubsection{Convolutional Neural Networks for ECG segmentation}

Building on our initial success in segmenting echos \cite{Zhang:wd}, we trained a convolutional neural network (CNN)-based model to delineate individual segments within the ECG. Convolutional neural networks and the broader approach of "deep learning" have revolutionalized the field of computer vision \cite{LeCun:2015dt}. Deep learning involves training a multilayer model, where each layer achieves an expanded representation of the layer below it. The lowest level takes in raw data and learn how to perform an initial level of abstraction, such as recognizing edges in an image. Each subsequent layer takes as input the layer below it and builds ever more complex abstractions. The top layer can then be used for a classification task, such as recognizing the subject of an image or localizing objects within it (i.e. segmentation). Unlike heuristic-based approaches, CNN models do not require user-specified rules but instead rely on abundant amounts of labeled input data. They then learn the rules needed to perform the desired task.

As training data, we downloaded raw ECG voltage data from two sources: 112 ECGs from the PTB Diagnostic database \cite{Goldberger:2000up} and 58 ECGs from the UCSF database.  For each ECG, we extracted a two-second strip and manually assigned to each one millisecond block one of six possible labels: P wave, PR segment (termination of P wave to start of QRS), QRS complex, ST segment, T wave, and TP segment.

We then trained a multilayered neural network to detect these segments within an ECG. The architecture of our network was based on the U-net network \cite{Ronneberger:2015vw} (\ref{fig:figure2} A). Our network accepted a 12 x 2000 input vector and was composed of sequential contracting and expanding paths with a total of 32 convolutional layers, 5 max pool layers, and 3 deconvolutional layers. The output of this CNN is a vector of ECG segment classes, identical in length to the input vector. (Supplementary Note \ref{note:Note1})

\begin{figure*}
\centering
\includegraphics[width=.8\linewidth]{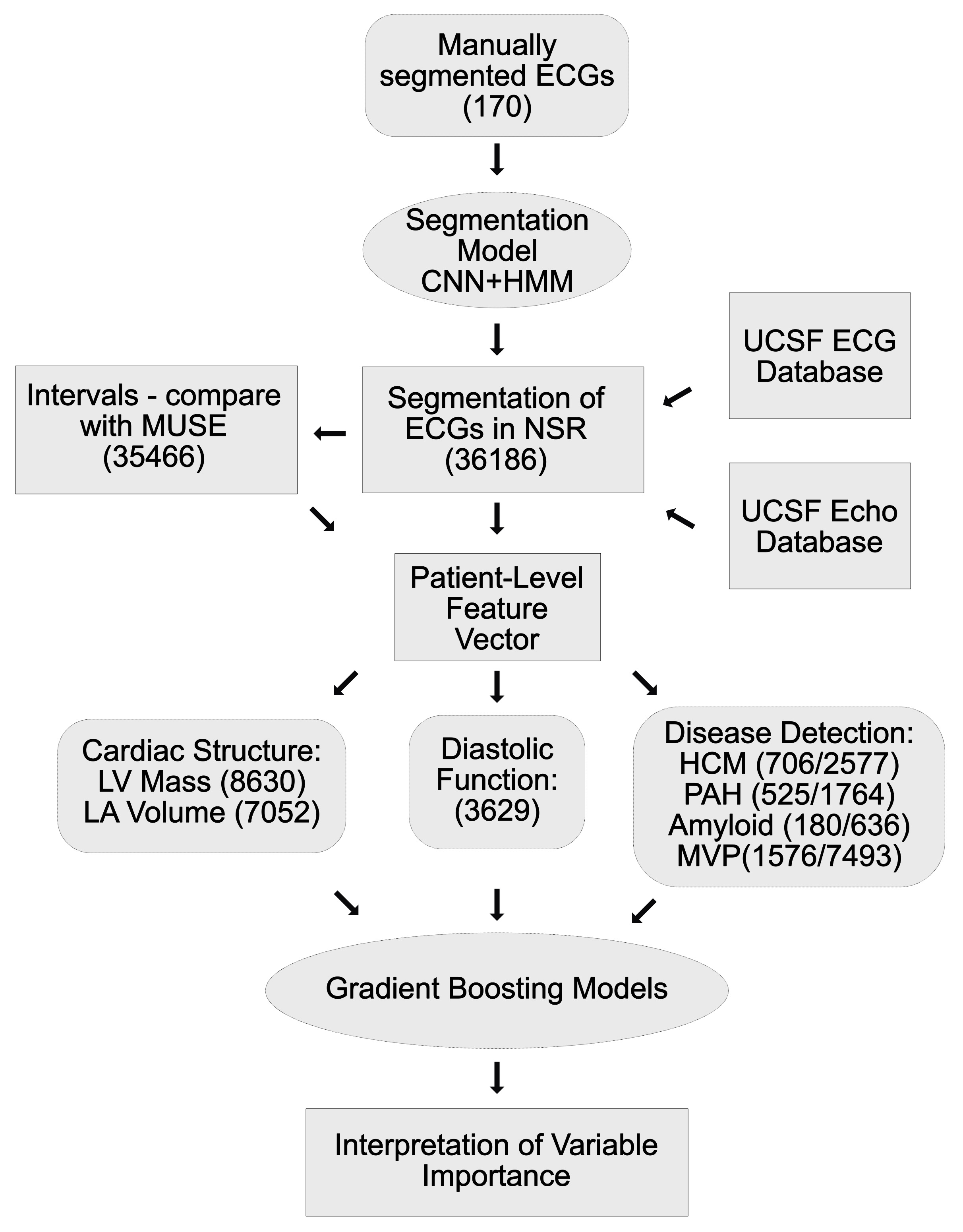}
\caption{\textbf{Workflow for ecgAI project. An ECG segmentation model was training using 170 manually labeled ECGs.} A subset of ECGs were selected from the UCSF for training interpretable models to estimate cardiac structure and function and detect and track disease. Segmentation of these ECGs enabled computation of standard physiologic intervals, which were then compared with the output of the MUSE/UCSF reference data. ECGs with good agreement were used to derive a 725-element patient-level ECG profile vector, which then served as input to train regression and classification models using the gradient boosting algorithm. The primary ECG features underlying each of these models was examined. Number of ECGs used for the various tasks are indicated in parentheses. For disease detection, a slash separates the number of cases and control ECGs used. Curved rectangles represent training data; ellipses represent algorithms; and standard rectangles represent other data types. CNN = Convolutional Neural Network; HMM = Hidden Markov Model; NSR = Normal Sinus Rhythm; HCM = Hypertrophic cardiomyopathy; PAH = Pulmonary arterial hypertension; MVP = Mitral valve prolapse; LV = left ventricle; LA = left atrium.}
\label{fig:figure1}
\end{figure*}

\subsubsection{Enhancing ECG segmentation with Hidden Markov Models}
Although U-Nets can provide accurate segmentation of objects, they fail to take advantage of the obligate ordering of elements in a typical ECG. For example the ST segment must follow a QRS complex and T waves should follow the ST segment. We thus trained a second machine learning model known as a Hidden Markov Model to accept the output of the U-Net and provide improved segmentation. Hidden Markov models (HMM) consists of four elements \cite{Rabiner:1989hs}: 1) an exhaustive set of different states, which in our case represent the different of segments of the ECG; 2) baseline probabilities of the states, reflecting the relative duration of the various segments; 3) the probability of moving from one state to another, which is captured in a  transition probability matrix; and 4) an emission probability matrix. The emission probability matrix describes the probability of seeing a given state in the input data, conditional on the true underlying state. It addresses the issue that noisy input data, such as the output of the U-Net, has many examples where the true state is incorrectly assigned. The probabilities of these errors are not uniform - for example noise in the TP segment is sometimes interpreted as a new P wave, but one is unlikely to confuse the TP segment with a QRS complex. To train the HMM, we input baseline and transition probabilities based on our manually segmented data and allowed the model to learn emission probabilities from the U-Net output on the training data.

As a final step, we introduced a series of simple heuristic filters to eliminate implausibly short ECG complexes (i.e. <10 mseconds).

\subsubsection{Validation of ECG segmentation}
Validation of segmentation performance was done in two ways. First we computed the Intersection over Union metric, or IoU, to compare the model output to manual labels. The IoU takes the number of pixels which overlap between the ground truth and automated segmentation (for a given class, such as the QRS complex) and divides them by the total number of pixels assigned to that class by either method. It ranges between 0 and 100. 

Second, we calculated standard ECG intervals based on the CNN-segmented ECGs for all ECGs in our sample and compared these against 141,864 ECG intervals derived from the GE MUSE software (these were derived from the 35,466 ECGs for which all four intervals were computed by both methods). Concordance between intervals was assessed using absolute differences, as a percentage of the reference (MUSE) value.

\subsection{Deriving patient-level ECG profiles}
Because ECG waveforms and intervals have corollaries to electrical and structural cardiac physiology, a crucial principle to our approach aimed to create a representation of the raw ECG data which preserves these features while still decreasing the feature space, making it tractable for analysis by interpretable machine learning algorithms. This approach also facilitates longitudinal tracking of clinically important features over time. To achieve this, we developed a 725-component ECG vector representation consisting of the following components, all of which were derived from corresponding segments of the CNN-segmented ECG. The PR interval, P-wave duration, QRS interval, heart rate, and QT intervals were calculated and averaged across all cardiac cycles and across 12-leads, and the five averaged values were included as five components in the ECG vector. For each of the following segments, the vector of raw-voltage amplitude from each of the 12 leads was resized to 20 pixels by linear interpolation, averaged across all cardiac cycles, and included as ECG vector components (totaling 720 components): the PR interval, the QRS complex and the ST-T-wave complex (including both the ST segment and the T wave). 

This 725-component ECG vector representation was calculated for each study ECG and input into machine learning algorithms, as below, for estimation of cardiac structure and function and for disease detection. Distinct from inputing raw ECG voltage data into a neural network \cite{Rajpurkar:2017wra, Yildirim:2018go} this vectorization process preserves meaningful representations of features within the ECG, facilitating interpretability. 

\subsection{ECG-Derived Estimates of Cardiac structure and Function}
The ECG patient vector was used as an input to train models to estimate left ventricular mass (indexed for body surface area, LVMi), left atrial volume (indexed, LAVOLi) and mitral annular medial e' (medial e'). Anticipating complex interactions among input features, as well as heterogeneity amongst patients \cite{Deo:2015hy}, we employed a machine learning algorithm known as a Gradient Boosted Machine \cite{Friedman:2001ue} (GBM), which is an ensemble regression-tree based technique. In this technique, a large number of decision trees are fit sequentially, with each successive tree being fit to the residuals of the prior tree, allowing each tree to become an expert in a subset of the data. In addition to being among the most powerful machine learning techniques for both classification and regression, the relative importance of input predictors in GBM models can be examined through variable importance analysis, providing results that are interpretable with respect to ECG representations of cardiac physiology. Individual GBM models were trained to estimate the three continuous structure and function metrics. We also generated dichotomous measures for each of these, treating controls as individuals with values below (for LVMi and LAVOLi) or above (medial e') the median value, and cases as individuals above or below the 10th percentile (Tables \ref{tab:tableS5}, \ref{tab:tableS6}, and \ref{tab:tableS7}). Given that we noted occasional inaccuracy in both our CNN-HMM segmentation model as well as in the MUSE values, we limited our models to ECGs with substantial agreement (mean difference < 10\%) across the RR, PR, QRS, and QT intervals. There was no appreciable difference in patient characteristics for this subset (Tables \ref{tab:tableS1}, \ref{tab:tableS2}, and \ref{tab:tableS3}). Models were fit using the GBM function in the R caret package. Tuning parameters were selected in an automated manner using 3-fold cross-validation. Accuracy was assessed using 5-fold cross validation, with AUROC curves used to evaluate classification tasks and absolute differences (50th, 75th, and 95th percentiles) and Bland-Altman plots \cite{Bland:1986je} used for continuous measures. Variable importance was extracted for each of the 725 features and averaged over cross-validation runs. To facilitate interpretation, values for voltage variation in ECG leads were binned so that each segment (e.g. QRS) was represented by 5 rather than 20 bins.

\subsection{Disease detection and tracking: Training Gradient Boosted Models to Quantify Diseases}
In addition to quantifying cardiac structure, we also trained GBM models using similar methods to detect PAH, HCM, CA, and MVP. Separate GBM models were trained to output a probability for each disease based on an input ECG vector. To demonstrate the use of this approach to track longitudinal changes in disease over time, we selected all patients who had ECGs in two or more years, and took the median score per patient for each year. Scores were plotted as a function of year.

\subsection{Statistical Methods}
All analyses were performed using R 3.3.2 or python 2.7. Differences between case and control characteristics for the diseases detection models were performed using two-tailed Wilcoxon-Mann-Whitney tests, t-tests, or chi-square tests. Only a single value was taken per patient in these pairwise comparisons.  The areas under the receiver operating characteristic curve for disease detection models were computed with the help of the pROC and hmeasure packages in R. Confidence intervals were generated by the method of Delong \cite{DeLong:1988gd}, as implemented in the pROC package. The only predictor for these models was the patient-level disease score, as output by the GBM model.

Convolutional neural networks were developed using the TensorFlow python package \cite{Abadi:CPnTLmNd}. Signal manipulation (such as linear interpolation for resizing) was performed using scikit-image \cite{vanderWalt:2014eo}.

\begin{figure*}
\centering
\includegraphics[width=.8\linewidth]{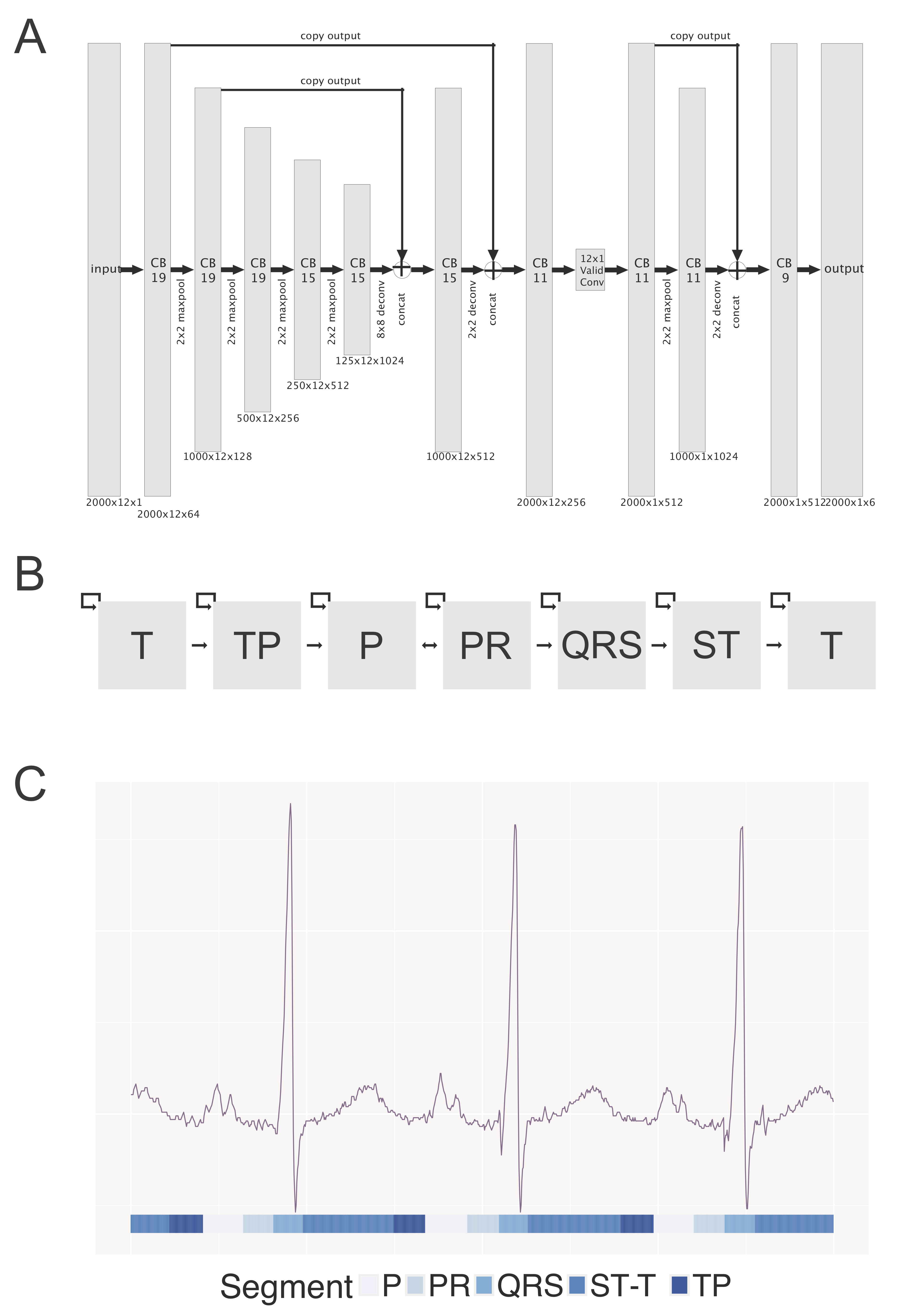}
\caption{\textbf{ecgAI method of ECG segmentation.}  A. Architecture of convolutional neural network used for ECG segmentation. Grey rectangles represent layers with dimensions listed below. The notation for each layer indicates the size of the input (e.g. 2000 ms, initially) by the number of leads (e.g. 12) by the number of filters. The size of the filter is specified in the body of the rectangle. B. Architecture of HMM used after CNN-based segmentation. Gray boxes represent "states" that are traversed in order in the ECG. Arrows represent transitions between time steps, which can result in remaining within a state, or making a transition to the next one. C. Example of CNN-HMM output for an ECG. CNN-HMM based classes are shown below the image. The ST and T wave segments have been combined.}
\label{fig:figure2}
\end{figure*}

\section*{Results}
\subsection{Validation of ecgAI Machine Learning-based ECG Segmentation}
Our ecgAI algorithmic pipeline (Figures \ref{fig:figure2}A and \ref{fig:figure2}B)  was trained on 170 manually segmented ECGs, and deployed on 36,186 sinus rhythm ECGs (Figure \ref{fig:figure1}). Example output from the ecgAI model is shown in Figure \ref{fig:figure2}C, with every time-step along the ECG tracing being classified as belonging to one of the six segments (illustrated in the Figure by separate colors). 

\begin{figure*}
\centering
\includegraphics[width=.8\linewidth]{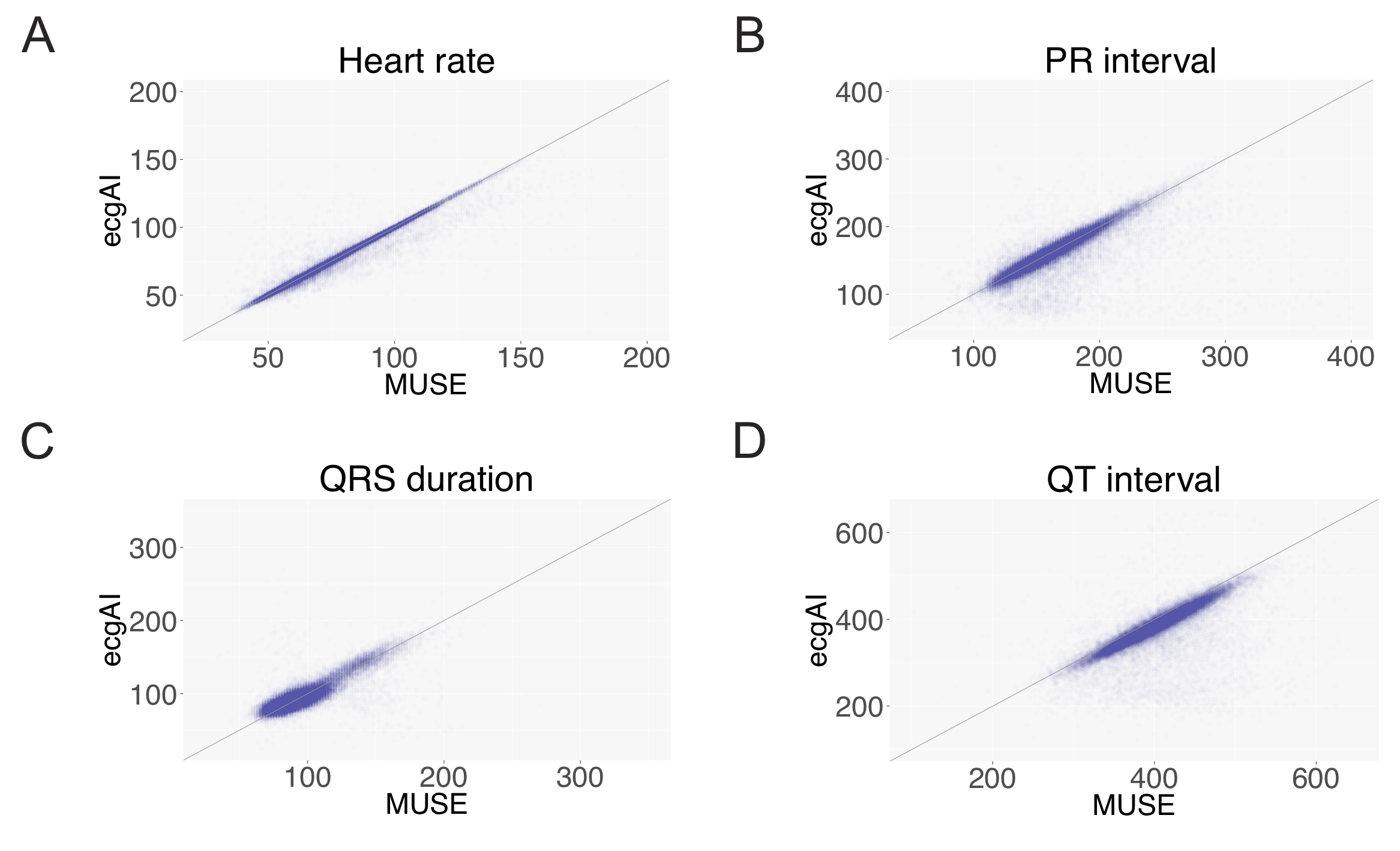}
\caption{\textbf{Comparison of ecgAI (HMM+CNN) derived measurements and MUSE/UCSF values for four commonly reported ECG measurements.} Each scatterplot depicts 35,466 comparisons. The line y=x is drawn to help identify any bias. The unit for heart rate is beats per minute while that of the other three metrics is milliseconds.}
\label{fig:figure3}
\end{figure*}

The IoU metrics for ECG segmentation were 91 (P wave), 85 (PR segment), 94 (QRS complex), 88 (ST segment), 91 (T wave), and 92 (TP segment). As a second indirect validation of segmentation performance, standard ECG interval measurements were calculated based on ecgAI segmentation on 35,466 ECGs not included in the training set and compared against the reference MUSE values (Table \ref{tab:table2}). Overall, intervals calculated from ecgAI-derived segmentation demonstrated good agreement with MUSE calculated intervals. Median absolute deviation between ecgAI-derived intervals was <6\% when compared to MUSE interpreted intervals, with Heart Rate, PR, QRS and QT intervals exhibiting 0.6\%, 3.0\%, 5.6\% and 4.4\% median absolute deviation, respectively (Table \ref{tab:table2}). Intervals from ecgAI-measurements demonstrated a strong correlation with those from MUSE ($\rho$=0.77-0.98, Figure \ref{fig:figure3}).

 \begin{table*}[] 
   \small 
   \centering 
   \begin{tabular}{| c | c | c | c | c | c |} 
   \hline
   \multirow{2}{*}{\textbf{Metric}} &
      \multirow{2}{*}{\parbox{4cm}{\centering \textbf{Number of ECGs Used for Comparison}}}&
      \multirow{2}{*}{\textbf{Median Value (IQR)}}&
      \multicolumn{3}{|c|}{\parbox{3cm}{\centering \textbf{Absolute Deviation: ecgAI vs. Reference (as \% of Reference)}  }} \\
      \cline{4-6}
    &  &  & 50$\%$ & 75$\%$ & 95$\%$\\
   \hline
   Heart Rate (beats/minute) & 35466 & 73 (63--86) & 0.6 & 1.5 & 9.8 \\\hline
   PR Interval (ms) & 35466 & 160 (144--180) & 3.0 & 5.9 & 27.4 \\\hline
   QRS Duration (ms) & 35466 & 90 (82--100) & 5.6 & 9.7 & 18.7 \\\hline
   QT Interval (ms) & 35466 & 402 (374--430) & 4.4 & 6.3 & 23.6 \\\hline
   Left ventricular mass index (g/m$^2$) & 8631 & 79.8 (66.2--97.3) & 16.5 & 28.9 & 59.8\\\hline
   Mitral annulus e' (cm/sec) & 3629 & 0.071(0.056--0.090) & 19.1 & 33.4 & 71.4 \\\hline
	Left atrial volume index (mL/m$^2$) & 7053 & 26.9 (20.9--35.3) & 22.9 & 41.2 & 99.1\\\hline
   \end{tabular}
   \caption[]{\textbf {Comparison of ecgAI-derived measurements and those  derived from MUSE (ECG inervals) or 2-dimensional echocardiography (structure/function metrics).} The absolute differences between ecgAI and reference values are reported as \% of reference measurements in order to compare across metrics. For each metric, 50$\%$, 75$\%$, and 95$\%$ of studies have an absolute difference between automated and manual measurements that is less than the value included in the corresponding columns. IQR = interquartile range.}
\label{tab:table2}
\end{table*}

\subsection{ecgAI Performance to Quantify Cardiac Structure}
\begin{figure*}
\centering
\includegraphics[width=.9\linewidth]{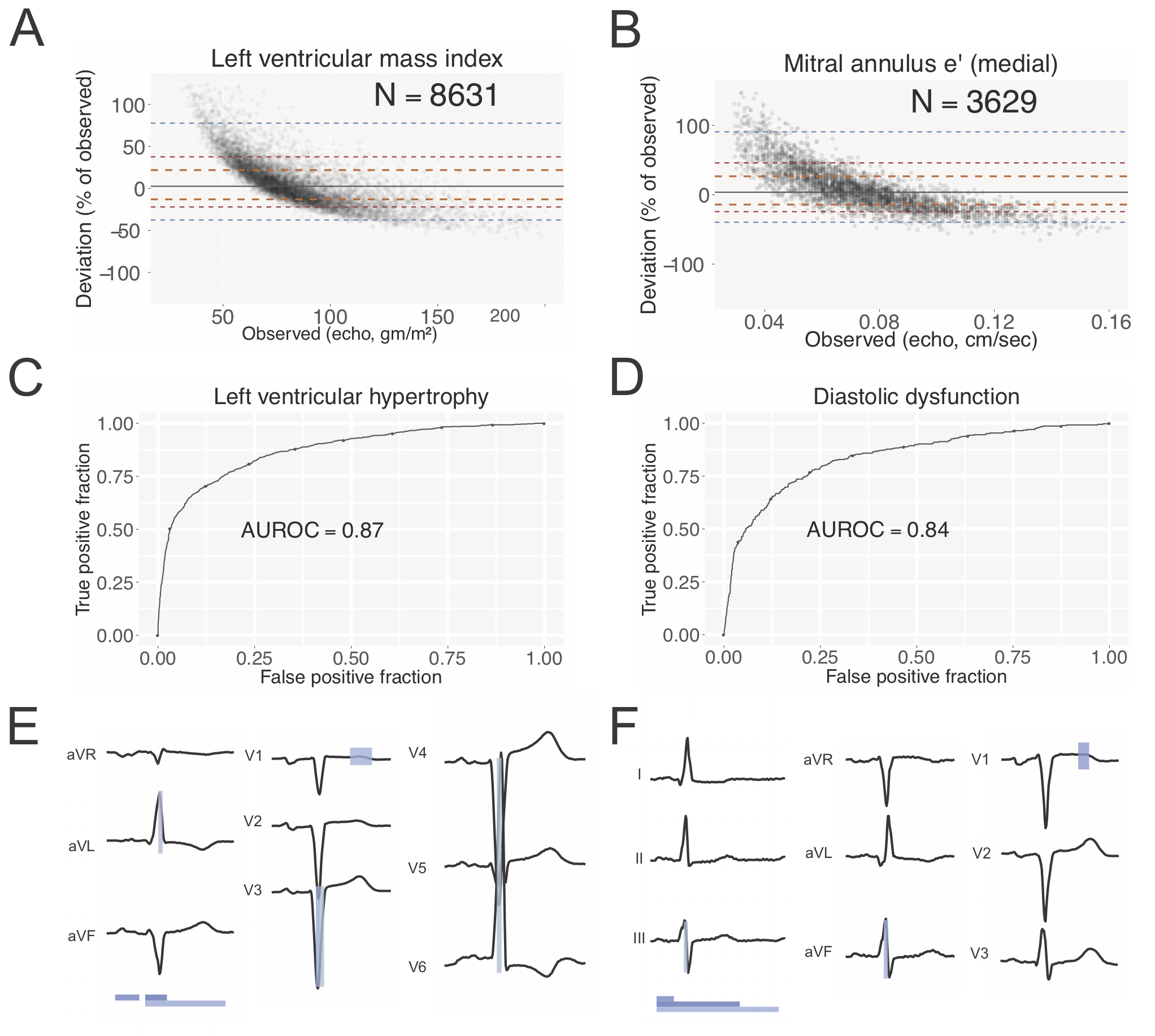}
\caption{\textbf{Estimating cardiac structure and function using patient-level ECG profiles.} Bland-Altman plots comparing estimates of left ventricular mass index (A) and mitral annulus medial e' (B) values using ECG alone compared to echo-derived values. Number of studies depicted in comparison is shown. Orange, red, and blue dashed lines delineate the central 50$\%$, 75$\%$ and 95$\%$ of patients, as judged by difference between automated and manual measurements.  The solid gray line indicates the median. Receiver operating characteristic (ROC) curves for classification models for left ventricular hypertrophy (C) and diastolic dysfunction (D). The area under the ROC curve is indicated. Variable importance for LVMi (E) and medial e' (F) estimation models. The predictors most important for each model are highlighted with the relative importance indicated by the shading (white to blue). Informative intervals are depicted below the plot while lead-specific segments of the ECG are highlighted on the voltage trace.}
\label{fig:figure4}
\end{figure*}

Though it is not standard to quantify the severity of cardiac structural abnormalities using ECGs, the presence of continuous measurements in the gold-standard echocardiographic studies enabled us to train ecgAI to estimate quantitative metrics. Median absolute deviation of ecgAI predictions against reference echo measurements varied by structure: the lowest deviation was for LVMi (16.5\%), intermediate deviation was for mitral annulus medial e' (19.1\%), and the greatest deviation was for LAVOLi (22.9\%) (Table \ref{tab:table2}). For all three structural measurements, there was a tendency to overestimate low values and underestimate high values (Figures \ref{fig:figure4}A, \ref{fig:figure4}B and \ref{fig:figureS1}), suggesting a more limited dynamic range for ECG compared to echo. When the continuous measurements for the cardiac structures were dichotomized, the model demonstrated strong discrimination for both left ventricular hypertrophy and diastolic dysfunction with AUROCs of 0.87 (95\% confidence interval: 0.86-0.89) and 0.84 (95\% CI 0.82-0.86) respectively (Figure \ref{fig:figure4}C, \ref{fig:figure4}D). Left atrial enlargement had a much lower AUROC of 0.62 (95\% CI 0.60-0.64), most likely reflecting a failure of the ECG to correctly estimate large atrial volumes. 

We identified those ECG components (waveform voltages and intervals from the 725-component patient-level ECG profile) which most strongly contributed to classification for each cardiac structural abnormality (Figure \ref{fig:figure4}E, F, Table \ref{tab:tableS11}. For LVMi, QRS duration was the strongest predictor with a variable score of 4.0, followed by P wave duration (3.3), QT duration (1.7), the middle portion of the QRS from lead V3 (1.5, segments 8-12 out of a total of 20) and the middle portion of the ST-T complex from lead V1 (1.3, segments 12-16) (Figure \ref{fig:figure4}E, Table \ref{tab:tableS11}). Collectively, these reflect many of the classic criteria for left ventricular hypertrophy \cite{Hancock:2009hu}.

For medial e' the strongest predictors were PR duration (3.1), QT duration (2.9), P wave duration (2.4), the middle portion of the ST-T complex from lead V1 (1.8, segments 8-12), and heart rate (1.2). For LAVOLi, top predictors were QT duration (4.6), P wave duration (4.5), QRS duration (1.4), PR duration (1.3) and the middle portion of the QRS from lead V6 (0.97).

\subsection{ecgAI Performance for Cardiac Disease Detection}

\begin{figure*}
\centering
\includegraphics[width=.9\linewidth]{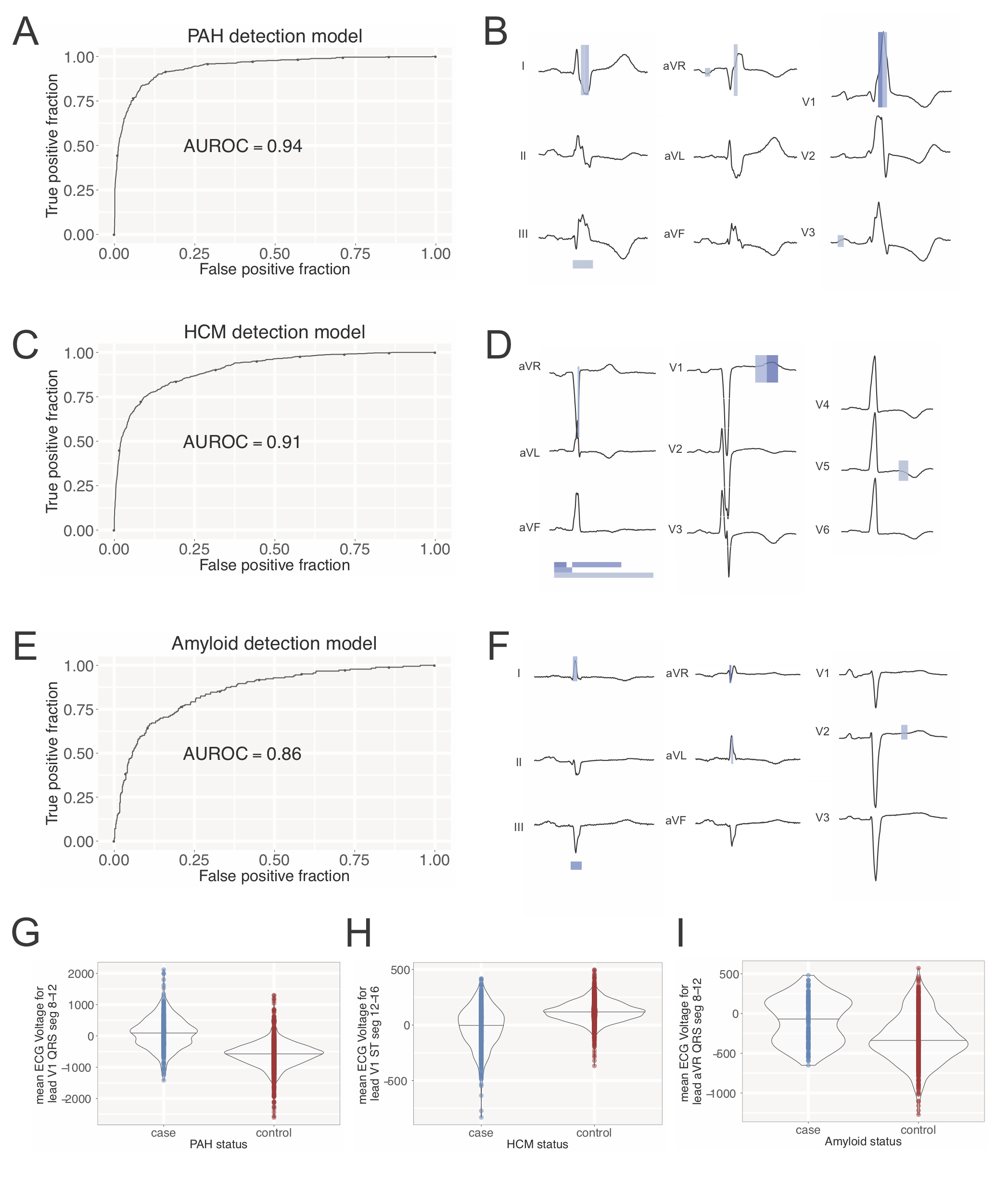}
\caption{\textbf{Detecting disease using patient-level ECG profiles.} ROC curves (with AUROC indicated) for disease detection models for PAH (A), HCM (C), CA (E), and MVP (F). Corresponding variable importance plots (B, D, F, H) with coloring as in  Figure \ref{fig:figure4}. Violin plots indicating distribution of the top predictive feature in cases and controls for PAH (G), HCM (H), and CA (I). All are significant at p<10$^{-6}$.}
\label{fig:figure5}
\end{figure*}
In addition to quantifying cardiac structure, we applied ecgAI toward disease classification and the discovery of ECG predictors of each disease. (Figure \ref{fig:figure5}).  The strongest discrimination was observed for a model for PAH which had an AUROC of 0.94 (95\% CI 0.93-0.95). Key predictors for PAH included the middle portion of QRS from lead V1 (variable score = 4.5, segments 8-12), reflecting a tall R' (Figure \ref{fig:figure5}G, p<2x10$^{-16}$), followed by the latter and middle portions of the QRS from lead V1 (1.6, segments 12-16; 1.4 segments 12-16), reflecting a deep S wave; and the early portion of the P-PR complex from lead V3 (0.9, segments 4-8) and aVR (0.9, segments 4-8), presumably reflecting right atrial enlargement (Figure \ref{fig:figure5}A-\ref{fig:figure5}B, Table \ref{tab:tableS12}).

HCM had the next strongest discrimination with an AUROC of 0.91 (95\% CI 0.90-0.92). The strongest predictors of HCM were the latter portion of the ST-T complex from lead V1 (3.8, segments 12-16), which can be markedly deeper in some HCM patients (Figure \ref{fig:figure5}H, p<2x10$^{-16}$), the P wave duration (3.5), QT duration (2.7), PR duration (2.4), and the middle portion of the QRS from lead aVR (1.3, segments 12-16) (Figure \ref{fig:figure5}C-\ref{fig:figure5}D, Table \ref{tab:tableS12}). 

CA had an AUROC of 0.86 (95\% CI 0.82-0.89), and the strongest predictors in this model were the early portion of the QRS from lead aVR (3.0, segments 4-8), which is blunted in voltage in CA patients ((Figure \ref{fig:figure5}I, p=3x10$^{-7}$), QRS duration (1.3), the middle and early portions of the QRS from lead I (1.2, segments 8-12; 1.1, segments 4-8), and the earliest portion of the QRS from lead V1 (1.1, segments 0-4) (Figure \ref{fig:figure5}E-\ref{fig:figure5}F, Table \ref{tab:tableS12}). 

The MVP showed the weakest discrimination, with an AUROC of 0.77 (95\% CI 0.76-0.78, \ref{fig:figure5}), a disease not known to strongly impact ECG morphology. The top predictors for MVP included PR duration (3.3), the early portion of the QRS from lead V2 (1.2, segments 4-8), the earliest portion of the QRS from lead V3 (1.2, segments 0-4), P wave duration (1.1) and QT duration (0.97) (Figure \ref{fig:figureS2}, Table \ref{tab:tableS12}).

\subsection{Serial ECGs Analysis with ecgAI to Perform Within-Patient Disease Tracking}

\begin{figure*}
\centering
\includegraphics[width=.9\linewidth]{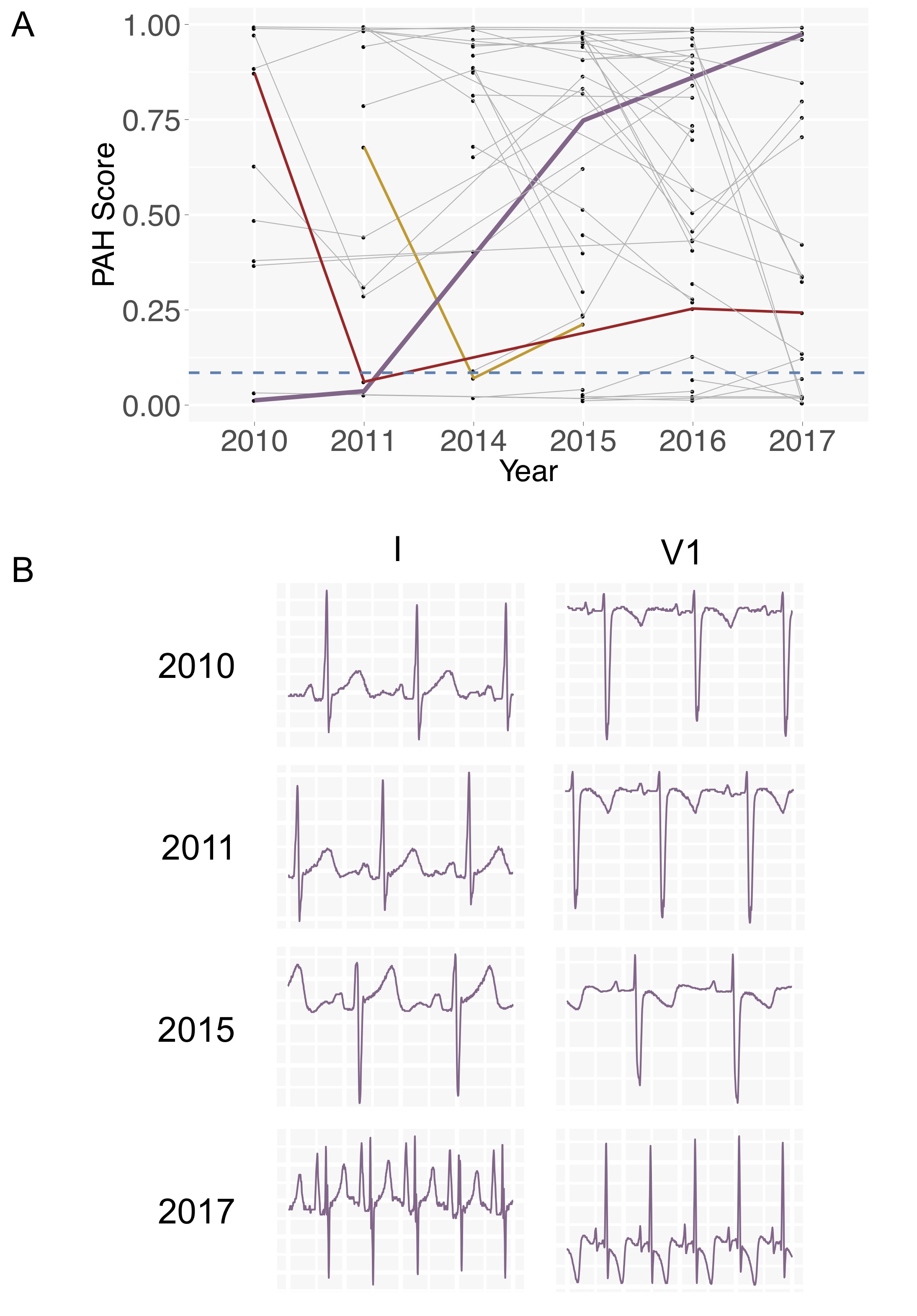}
\caption{\textbf{ECG-profile based models can be used to track changes in patient ECGs in PAH.} A. Scores for PAH detection for individual patients with measurements for 2 or more years. A median of all scores for each year is computed and lines are drawn connecting scores for different years for each patient. The blue dashed line indicates a score threshold with 90\% specificity and 80\% sensitivity for a diagnosis of PAH. Purple, red and yellow lines highlight score trajectories for three patients with dramatic variation in scores, crossing this threshold. B. Variation in ECG patterns for leads I and V1 from 2010-2017 for patient highlighted in purple in A. Over time, as the PAH score increases, the QRS axis swings progressively rightward (lead I), the P wave height grows, and the R' wave in lead V1 increases in size.}
\label{fig:figure6}
\end{figure*}

\begin{figure*}
\centering
\includegraphics[width=.9\linewidth]{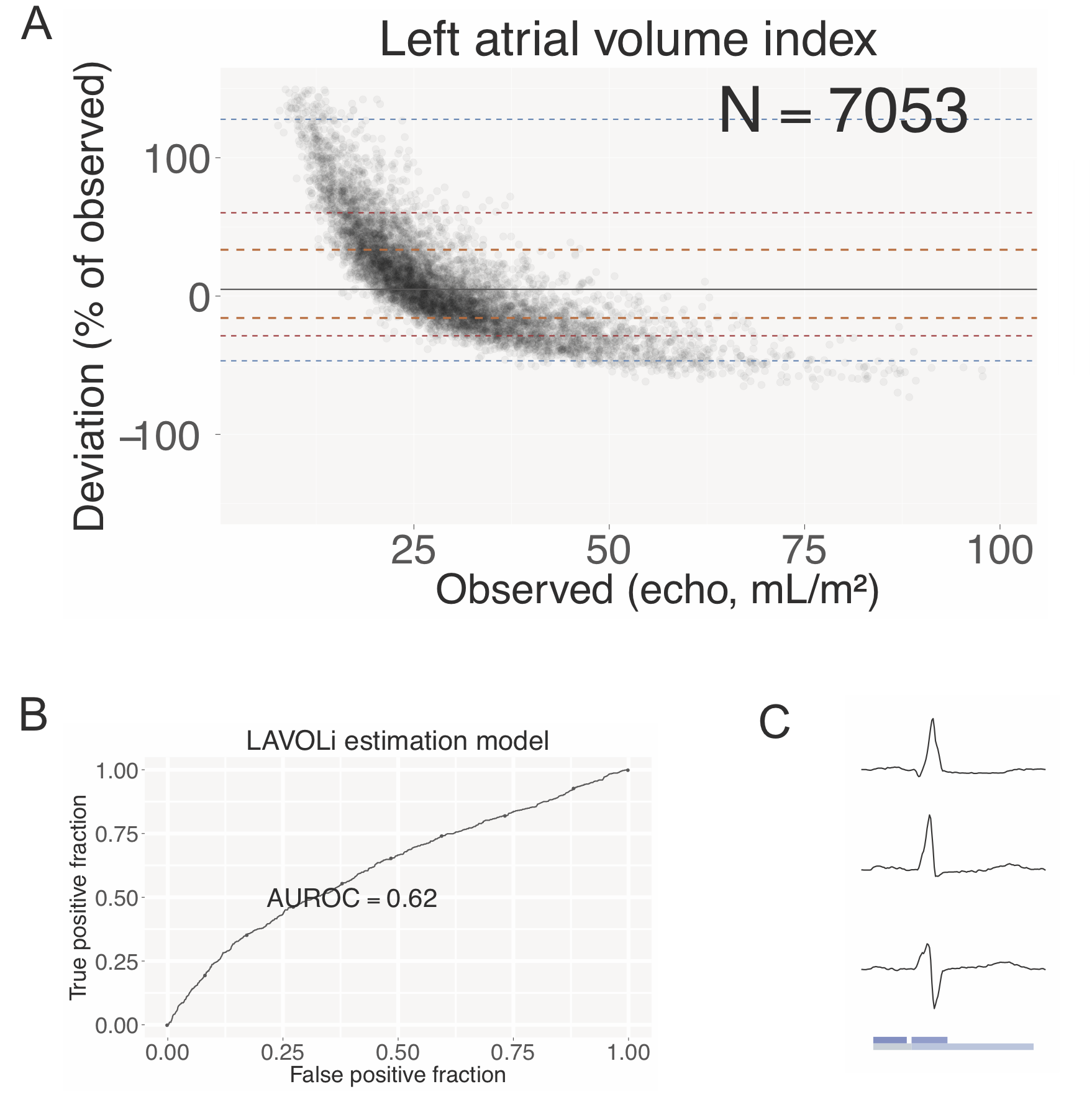}
\caption{\textbf{Estimating left atrial volume index using patient-level ECG profiles.} A. Bland-Altman plot comparing estimation of LAVOLi values using ECG alone compared to echo-derived values. B. ROC curve for a binary classification of left atrial enlargement, as defined by being in the upper 90th percentile of indexed left atrial volume. C. Variable importance as in Figure \ref{fig:figure4}}
\label{fig:figureS1}
\end{figure*}

\begin{figure*}
\centering
\includegraphics[width=.9\linewidth]{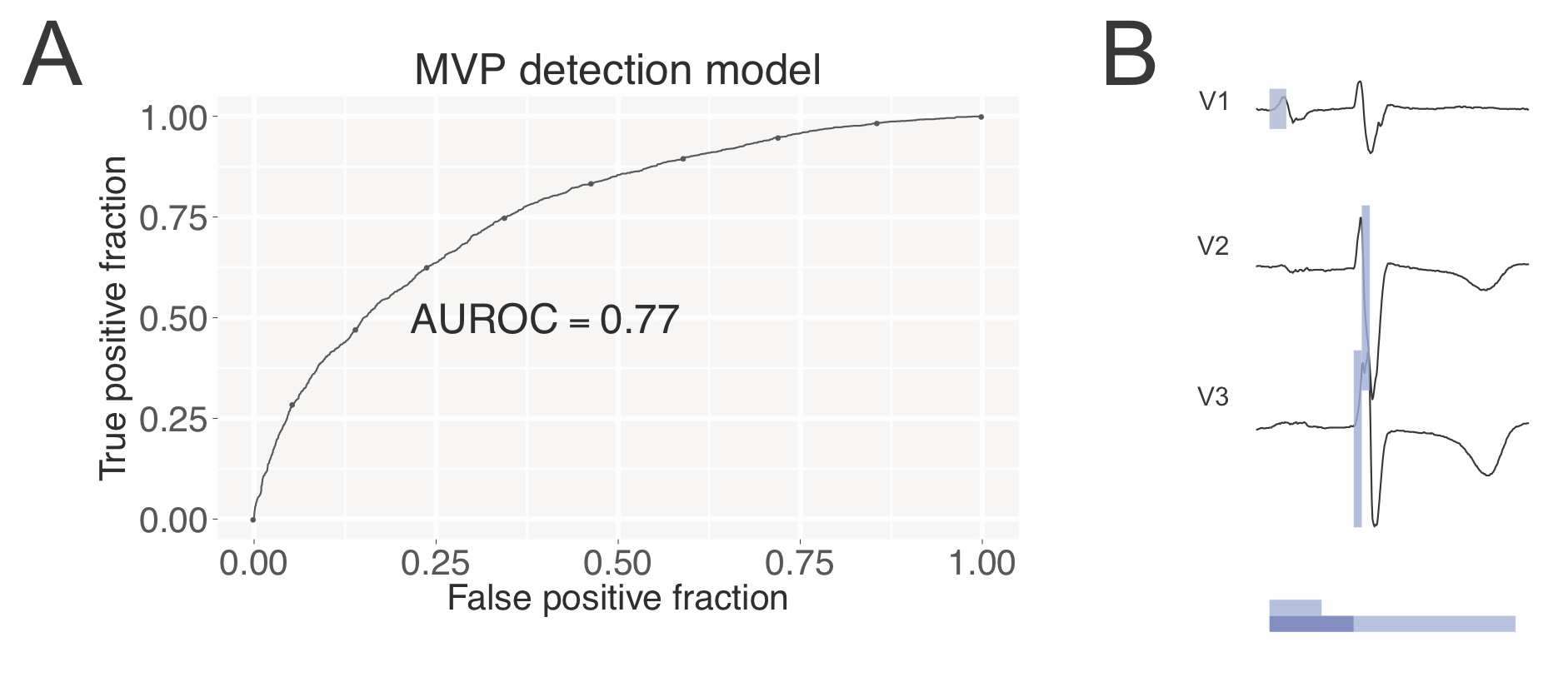}
\caption{\textbf{Detecting disease using patient-level ECG profiles.} ROC curves (with AUROC indicated) for disease detection models for  MVP (A). Corresponding variable importance plots (B) with coloring as in Figure \ref{fig:figure4}}
\label{fig:figureS2}
\end{figure*}

By applying ecgAI to serial ECGs of PAH patients, we obtained a progression of scores over time corresponding to the degree to which the model estimated likelihood of PAH based on ECG features (Figure \ref{fig:figure6}A). The dashed blue line represents the PAH score at which PAH is identified with 80\% sensitivity and 90\% specificity. Patients typically have scores that remain with a narrow range but there are some exceptions - and we highlight the three most prominent ones.  Figure \ref{fig:figure6}B shows a time course of ECG tracings for the individual depicted by the purple trajectory in Figure \ref{fig:figure6}A). In 2010 and 2011, ECG tracings do not have any marked abnormalities. In 2015 and 2017, ECG tracings appear increasingly abnormal, with a prominent R wave and T wave inversion in lead V1, and QRS changes and a tall prominent P wave in lead I. These progressive ECG changes over time correspond with the increasing PAH scores from 2010-2017. 

\begin{figure*}
\centering
\includegraphics[width=.7\linewidth]{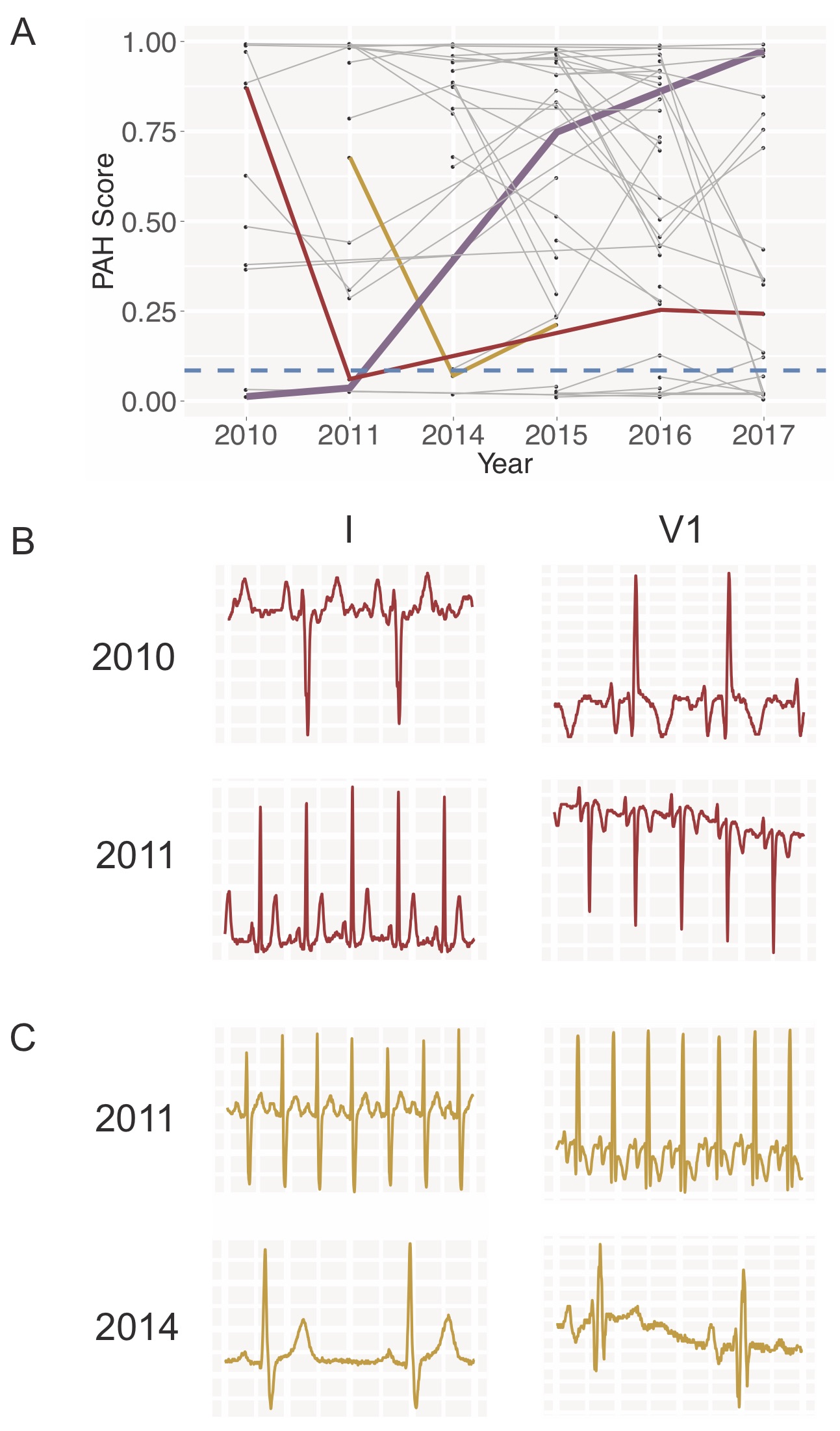}
\caption{\textbf{ECG-profile based models can be used to track changes in patient ECGs in PAH.} A. Scores for PAH detection for individual patients with measurements for 2 or more years (repeat of Figure \ref{fig:figure6}A). A median of all scores for each year is computed and lines are drawn connecting scores for different years for each patient. The blue dashed line indicates a score threshold with 90\% specificity and 80\% sensitivity for a diagnosis of PAH. Purple, red and yellow lines highlight score trajectories for three patients with dramatic variation in scores, crossing this threshold. B. Variation in ECG patterns for leads I and V1 for patients highlighted in red (B) and yellow (C) in A. In each case, ECGs show substantial temporal changes in complex morphology, particularly in those features important for a PAH diagnosis (Figure \ref{fig:figure5})}
\label{fig:figureS3}
\end{figure*}

Two other patients (trajectories colored in red and yellow) had precipitous decreases followed by subsequent increases in score (Figure \ref{fig:figureS3}).  In both cases, the ECG tracings from the high PAH score year appear abnormal, featuring prominent R waves in V1 and a more negatively directed QRS vector in lead I. In contrast (and for unclear reasons), the subsequent low PAH score ECG tracings for both individuals appear substantially different and more normal, with a decrease in R wave prominence in V1 and normalization of the QRS in lead I (Figure \ref{fig:figureS3}B). The GBM PAH score thus tracks well with visible morphological change in the ECG.

\section*{Discussion}
In keeping with a widespread adoption of machine learning across nearly every industry, there has been a dramatic increase in publications applying these methods to carry out routine diagnostic tasks in medicine. Most of these have emphasized matching or even outperforming practicing physicians, whether it be for interpreting retinograms \cite{Gulshan:2016iu}, skin disorders \cite{Esteva:2017ct}, chest x-rays \cite{Rajpurkar:tu}, mammograms \cite{Geras:2017um}, bone x-rays \cite{Rajpurkar:2017tg}, heart rhythm abnormalities \cite{Rajpurkar:2017wra}, or deciphering which view was collected by a cardiac sonographer \cite{Gao:2017fk, Madani:2018hm}. As the field matures, it will be important to think carefully about how exactly these automated interpretation models will be accommodated within the current clinical workflow. Our current work, similar to our prior work on echocardiography \cite{Zhang:wd}, proposes how fully automated interpretation might enable studies that otherwise would not be done, such as detecting and tracking adverse cardiac remodeling in asymptomatic patients in a primary care clinic. Nonetheless, for all these applications that hope to guide medical decisions, an important question is whether the machine learning algorithms can provide physicians and patients an adequate explanation as to why a certain automated diagnosis or recommendation was made \cite{Lipton:2016tw}. Although in some examples that may be unnecessary - and a black box approach might suffice - in most cases where there is need of a complex decision, it is expected that a clear rationale is provided, ideally one which can be verified (in our case, visually) by the physician and potentially even by the patient.

Here we emphasize three facets of an artificial intelligence approach to ECG interpretation that differs from these prior works: 1) the use of machine learning to extend the utility of a diagnostic tool to applications beyond what would be possible by human readers; 2) the focus on eliciting interpretable features which can be used to both justify an automated diagnosis within clinical care and inspire new research on physiological correlates of disease; and 3) the demonstration of a flexible framework that permits estimation or classification for a broad range of cardiac metrics and diseases.  Although feature extraction has long been part of the typical ECG analysis pipeline \cite{Lyon:2018cv}, it has been performed primarily for the purpose of reducing the ECG to a simpler set of descriptors, since it was not tractable to input raw ECG data into classifier algorithms and yet maintain algorithm performance. The recent emergence of deep neural networks for ECG analysis \cite{Rajpurkar:2017wra,Hong:2017dc,Xiong:2017bb, Acharya:2018jq} arguably lessened both of these needs, making it possible to input raw ECG data and demonstrating performance gains for certain tasks. But neural networks, at least presently, are used at the cost of interpretability. Feature extraction within our framework aims instead to delineate portions of the ECG signal that have clinical meaning and culminates in the patient-level ECG profile, enabling the use of down-sampled raw data for machine learning. This results in outputs that are more interpretable, even compared to approaches where simple abstractions of the ECG signal, such as slopes, ratios, or displacement, are performed \cite{Hancock:2009hu}. While we apply this approach to three structural and four disease entities, this framework can be expanded broadly to any application, many of which may not be readily apparent. Notably, the composition of the ECG feature vector could also be modified in future applications to incorporate disease-specific knowledge in order to better capture specific aspects of cardiac physiology.

We believe that the enormous potential of applying machine learning to medicine must lie in its ability to illuminate patterns across large quantities of data in a way that preserves clinical interpretability, both to maintain physician and patient agency in decision-making and to enable knowledge discovery. And we suggest that this does not necessarily have to come at the cost of algorithm performance. In the case of ECG-disease correlates, there is ample evidence that previously recognized ECG-predictors represent only a fraction of informative features of any disease \cite{Hong:2017dc,Nahar:2013fh}, making the case for data-driven discovery of novel ECG correlates. Applying our analysis framework to detect several diseases demonstrated strong discriminative ability to identify PAH (AUROC=0.94) and HCM (AUROC=0.91), and slightly weaker ability for CA (AUROC=0.86) and MVP (AUROC=0.77), using ECG inputs alone. Because the ECG is only one component upon which the clinical diagnosis for any of these diseases is made, most prior studies tend to highlight the association of various ECG features with disease status, rather than describing the global discrimination performance \cite{Cheng:2012dt,Dumont:2006ee,Rahman:2015jw,Lyon:2018bk,Fikrle:2013de, Kopec:2012bp, Peighambari:2014ib} — limiting our ability to directly compare our performance. In work bearing the most similarity to ours \cite{Rahman:2015jw}, the authors used various ECG features to identify HCM patients from non-HCM patients, and also reported strong performance across several global metrics for HCM detection. The focus of their approach, however, remained the optimization of predictive performance rather than enabling clinical interpretability through informative ECG features. Similarly, we emphasize the distinction between our approach and others which use manually-derived features or proprietary software, which significantly limits scalability and interpretability \cite{Sengupta:2018hl}. In our study, the ECG-based features identified as most strongly contributing to prediction for each disease have clear physiologic parallels — such as ECG-correlates of right ventricular hypertrophy in PAH and myocardial infiltration in CA — which conforms to our expectations based on pathophysiology, and increases confidence in and acceptance of model performance. Furthermore, the novel predictors identified by our models may provide inroads into future investigation.

It has long been recognized that serial ECGs can reflect changes in cardiac structure and morphology, which can also correlate with risk for adverse outcomes \cite{Levy:1994ue}. Prior efforts have illustrated that ECG features change with disease progression \cite{Tonelli:2013ii, Henkens:2008ky} and even in response to treatment \cite{Waligora:2018ig}. Our approach is well suited to perform algorithmic longitudinal ECG tracking, and has the advantage of not being limited to \textit{a priori} derived ECG features, and instead can learn high level interactions or continuous feature weights from the data.

Although we used a specific machine learning pipeline to rapidly segmenting ECG in this work, other methods, including any of the existing heuristic based segmentation algorithms, could also be used to derive patient-level ECG profiles \cite{Laguna:1994up}. We also note as an additional limitation that our models are currently optimized to analyze ECGs in normal sinus rhythm. To obtain ECG profiles for patients with more complex rhythms (such as the presence of premature or paced beats) will require expanding our training data and the states considered in our HMM, though we envisage these would be straightforward to train within a CNN-HMM framework. Furthermore, although large in scale, our data still derived from a single medical center.

Quantitative patient tracking using the output of multidimensional models is not performed routinely for ECGs or even echos, in part because of long-standing fears that it might obscure the diagnostic process \cite{Brohet:1984ch,Kors:1990wn}. With the current wide-spread availability of digital data, we strongly believe such concerns should be revisited, both for the benefit of the physician and patient. To this end, a primary motivation of this work is to demonstrate how we can extract much more knowledge from our current low-cost input data, all in an automated manner, and yet remain transparent to physicians, patients, and researchers about the provenance of these insights.

\begin{acknowledgements}
We would like to acknowledge the assistance of Edward Mahoney and John Cooper in accessing records from the MUSE server. We would also like to thank Dr. Pulkit Agrawal for advice on the ECG segmentation model.
\end{acknowledgements}

\section*{Bibliography}
\bibliography{echo}

\begin{thebibliography}{54}
\providecommand{\natexlab}[1]{#1}
\providecommand{\url}[1]{\texttt{#1}}
\expandafter\ifx\csname urlstyle\endcsname\relax
  \providecommand{\doi}[1]{doi: #1}\else
  \providecommand{\doi}{doi: \begingroup \urlstyle{rm}\Url}\fi

\bibitem[Drazen et~al.(1988)Drazen, Mann, Borun, Laks, and
  Bersen]{Drazen:1988ud}
E~Drazen, N~Mann, R~Borun, M~Laks, and A~Bersen.
\newblock {Survey of computer-assisted electrocardiography in the United
  States.}
\newblock \emph{Journal of Electrocardiology}, 21 Suppl:\penalty0 S98--104,
  1988.

\bibitem[Bhatia et~al.(2017)Bhatia, Bouck, Ivers, Mecredy, Singh, Pendrith, Ko,
  Martin, Wijeysundera, Tu, Wilson, Wintemute, Dorian, Tepper, Austin, Glazier,
  and Levinson]{Bhatia:2017ji}
R~Sacha Bhatia, Zachary Bouck, Noah~M Ivers, Graham Mecredy, Jasjit Singh,
  Ciara Pendrith, Dennis~T Ko, Danielle Martin, Harindra~C Wijeysundera, Jack~V
  Tu, Lynn Wilson, Kimberly Wintemute, Paul Dorian, Joshua Tepper, Peter~C
  Austin, Richard~H Glazier, and Wendy Levinson.
\newblock {Electrocardiograms in Low-Risk Patients Undergoing an Annual Health
  Examination}.
\newblock \emph{JAMA Internal Medicine}, 177\penalty0 (9):\penalty0 1326--8,
  September 2017.

\bibitem[Pitts et~al.(2008)Pitts, Niska, Xu, and Burt]{Pitts:2008vw}
Stephen~R Pitts, Richard~W Niska, Jianmin Xu, and Catharine~W Burt.
\newblock {National Hospital Ambulatory Medical Care Survey: 2006 emergency
  department summary.}
\newblock \emph{National health statistics reports}, 7:\penalty0 1--38, August
  2008.

\bibitem[Blackburn et~al.(1960)Blackburn, Keys, Simonson, Rautaharju, and
  Punsar]{BLACKBURN:1960un}
H~Blackburn, A~Keys, E~Simonson, P~Rautaharju, and S~Punsar.
\newblock {The Electrocardiogram in Population Studies }.
\newblock \emph{Circulation}, 21:\penalty0 1160--1175, June 1960.

\bibitem[Rautaharju(2016)]{PhD:2016et}
Pentti~M Rautaharju.
\newblock {Eyewitness to history: Landmarks in the development of computerized
  electrocardiography}.
\newblock \emph{Journal of Electrocardiology}, 49\penalty0 (1):\penalty0 1--6,
  January 2016.

\bibitem[Kligfield et~al.(2007)Kligfield, Gettes, Bailey, Childers, Deal,
  Hancock, van Herpen, Kors, Macfarlane, Mirvis, Pahlm, Rautaharju, and
  Wagner]{Kligfield:2007ir}
P~Kligfield, L~S Gettes, J~J Bailey, R~Childers, B~J Deal, E~W Hancock, G~van
  Herpen, J~A Kors, P~Macfarlane, D~M Mirvis, O~Pahlm, P~Rautaharju, and G~S
  Wagner.
\newblock {Recommendations for the Standardization and Interpretation of the
  Electrocardiogram. Part I: The Electrocardiogram and Its Technology. A
  Scientific Statement From the American Heart Association Electrocardiography
  and Arrhythmias Committee, Council on Clinical Cardiology; the American
  College of Cardiology Foundation; and the Heart Rhythm Society. Endorsed by
  the International Society for Computerized Electrocardiology}.
\newblock \emph{Circulation}, pages 1--20, February 2007.

\bibitem[Casale et~al.(1985)Casale, Devereux, Kligfield, Eisenberg, Miller,
  Chaudhary, and Phillips]{MD:1985ck}
P~N Casale, R~B Devereux, P~Kligfield, R~R Eisenberg, D~H Miller, B~S
  Chaudhary, and M~C Phillips.
\newblock {Electrocardiographic detection of left ventricular hypertrophy:
  Development and prospective validation of improved criteria}.
\newblock \emph{JAC}, 6\penalty0 (3):\penalty0 572--580, September 1985.

\bibitem[Krizhevsky et~al.(2012)Krizhevsky, Sutskever, and
  Hinton]{Krizhevsky:2012wl}
A~Krizhevsky, I~Sutskever, and G~E Hinton.
\newblock {Imagenet classification with deep convolutional neural networks}.
\newblock \emph{Advances in neural information processing}, pages 1097--1105,
  2012.

\bibitem[van~den Oord et~al.(2016)van~den Oord, Dieleman, Zen, Simonyan,
  Vinyals, Graves, Kalchbrenner, Senior, and Kavukcuoglu]{vandenOord:2016uo}
Aaron van~den Oord, Sander Dieleman, Heiga Zen, Karen Simonyan, Oriol Vinyals,
  Alex Graves, Nal Kalchbrenner, Andrew Senior, and Koray Kavukcuoglu.
\newblock {WaveNet: A Generative Model for Raw Audio}.
\newblock \emph{arXiv.org}, September 2016.

\bibitem[Gulshan et~al.(2016)Gulshan, Peng, Coram, Stumpe, Wu, Narayanaswamy,
  Venugopalan, Widner, Madams, Cuadros, Kim, Raman, Nelson, Mega, and
  Webster]{Gulshan:2016iu}
Varun Gulshan, Lily Peng, Marc Coram, Martin~C Stumpe, Derek Wu, Arunachalam
  Narayanaswamy, Subhashini Venugopalan, Kasumi Widner, Tom Madams, Jorge
  Cuadros, Ramasamy Kim, Rajiv Raman, Philip~C Nelson, Jessica~L Mega, and
  Dale~R Webster.
\newblock {Development and Validation of a Deep Learning Algorithm for
  Detection of Diabetic Retinopathy in Retinal Fundus Photographs.}
\newblock \emph{Jama}, 316\penalty0 (22):\penalty0 2402--2410, December 2016.

\bibitem[Esteva et~al.(2017)Esteva, Kuprel, Novoa, Ko, Swetter, Blau, and
  Thrun]{Esteva:2017ct}
Andre Esteva, Brett Kuprel, Roberto~A Novoa, Justin Ko, Susan~M Swetter,
  Helen~M Blau, and Sebastian Thrun.
\newblock {Dermatologist-level classification of skin cancer with deep neural
  networks}.
\newblock \emph{Nature}, 542\penalty0 (7639):\penalty0 115--118, February 2017.

\bibitem[Lipton(2016)]{Lipton:2016tw}
Zachary~C Lipton.
\newblock {The Mythos of Model Interpretability}.
\newblock \emph{arXiv:1606.03490[cs.LG]}, June 2016.

\bibitem[Zhang et~al.(2018)Zhang, Gajjala, Agrawal, Tison, Hallock,
  Beussink-Nelson, Fan, Aras, Jordan, Fleischmann, Melisko, Qasim, Shah,
  Bajcsy, and Deo]{Zhang:wd}
J~Zhang, S~Gajjala, P~Agrawal, GH~Tison, LA~Hallock, L~Beussink-Nelson, E~Fan,
  MA~Aras, C~Jordan, KE~Fleischmann, M~Melisko, A~Qasim, SJ~Shah, R~Bajcsy, and
  RC~Deo.
\newblock A computer vision pipeline for automated determination of cardiac
  structure and function and detection of disease by two-dimensional
  echocardiography.
\newblock \emph{arxiv:1706.07342[cs.CV]}, January 2018.

\bibitem[Gersh et~al.(2011)Gersh, Maron, Bonow, Dearani, Fifer, Link, Naidu,
  Nishimura, Ommen, Rakowski, Seidman, Towbin, Udelson, and
  Yancy]{Gersh:2011goa}
Bernard~J Gersh, Barry~J Maron, Robert~O Bonow, Joseph~A Dearani, Michael~A
  Fifer, Mark~S Link, Srihari~S Naidu, Rick~A Nishimura, Steve~R Ommen, Harry
  Rakowski, Christine~E Seidman, Jeffrey~A Towbin, James~E Udelson, and Clyde~W
  Yancy.
\newblock {2011 ACCF/AHA Guideline for the Diagnosis and Treatment of
  Hypertrophic Cardiomyopathy A Report of the American College of Cardiology
  Foundation/American Heart Association Task Force on Practice Guidelines
  Developed in Collaboration With the American Association for Thoracic
  Surgery, American Society of Echocardiography, American Society of Nuclear
  Cardiology, Heart Failure Society of America, Heart Rhythm Society, Society
  for Cardiovascular Angiography and Interventions, and Society of Thoracic
  Surgeons}.
\newblock \emph{Journal of the American College of Cardiology}, 58\penalty0
  (25):\penalty0 e212--e260, December 2011.

\bibitem[Martinez et~al.(2004)Martinez, Almeida, Olmos, Rocha, and
  Laguna]{Martinez:2004ge}
J~P Martinez, R~Almeida, S~Olmos, A~P Rocha, and P~Laguna.
\newblock {A Wavelet-Based ECG Delineator: Evaluation on Standard Databases}.
\newblock \emph{IEEE transactions on bio-medical engineering}, 51\penalty0
  (4):\penalty0 570--581, April 2004.

\bibitem[LeCun et~al.(2015)LeCun, Bengio, and Hinton]{LeCun:2015dt}
Yann LeCun, Yoshua Bengio, and Geoffrey Hinton.
\newblock {Deep learning}.
\newblock \emph{Nature}, 521\penalty0 (7553):\penalty0 436--444, May 2015.

\bibitem[Goldberger et~al.(2000)Goldberger, Amaral, Glass, Hausdorff, Ivanov,
  Mark, Mietus, Moody, Peng, and Stanley]{Goldberger:2000up}
A~L Goldberger, L~A Amaral, L~Glass, J~M Hausdorff, P~C Ivanov, R~G Mark, J~E
  Mietus, G~B Moody, C~K Peng, and H~E Stanley.
\newblock {PhysioBank, PhysioToolkit, and PhysioNet: components of a new
  research resource for complex physiologic signals.}
\newblock \emph{Circulation}, 101\penalty0 (23):\penalty0 E215--20, June 2000.

\bibitem[Ronneberger et~al.(2015)Ronneberger, Fischer, and
  Brox]{Ronneberger:2015vw}
O.~Ronneberger, P.~Fischer, and T.~Brox.
\newblock U-net: Convolutional networks for biomedical image segmentation.
\newblock \emph{arXiv:1706.07342[cs.CV]}, May 2015.

\bibitem[Rabiner(1989)]{Rabiner:1989hs}
Lawrence~R Rabiner.
\newblock {A tutorial on hidden Markov models and selected applications in
  speech recognition}.
\newblock \emph{Proceedings of the IEEE}, 77\penalty0 (2):\penalty0 257--286,
  February 1989.

\bibitem[Rajpurkar et~al.(2017{\natexlab{a}})Rajpurkar, Irvin, Zhu, Yang,
  Mehta, Duan, Ding, Bagul, Langlotz, Shpanskaya, Lungren, and
  Ng]{Rajpurkar:2017wra}
Pranav Rajpurkar, Jeremy Irvin, Kaylie Zhu, Brandon Yang, Hershel Mehta, Tony
  Duan, Daisy Ding, Aarti Bagul, Curtis Langlotz, Katie Shpanskaya, Matthew~P.
  Lungren, and Andrew~Y. Ng.
\newblock {Cardiologist-Level Arrhythmia Detection with Convolutional Neural
  Networks}.
\newblock \emph{arXiv:1707.01836[cs.CV]}, July 2017{\natexlab{a}}.

\bibitem[Yildirim(2018)]{Yildirim:2018go}
{\"O}zal Yildirim.
\newblock {A novel wavelet sequence based on deep bidirectional LSTM network
  model for ECG signal classification}.
\newblock \emph{Computers in Biology and Medicine}, 96:\penalty0 189--202, May
  2018.

\bibitem[Deo(2015)]{Deo:2015hy}
Rahul~C Deo.
\newblock {Machine Learning in Medicine.}
\newblock \emph{Circulation}, 132\penalty0 (20):\penalty0 1920--1930, November
  2015.

\bibitem[Friedman(2001)]{Friedman:2001ue}
J~H Friedman.
\newblock {Greedy function approximation: A gradient boosting machine}.
\newblock \emph{The Annals of Statistics}, 29\penalty0 (5):\penalty0
  1189--1232, October 2001.

\bibitem[Bland and Altman(1986)]{Bland:1986je}
J~M Bland and D~G Altman.
\newblock {Statistical Methods for Assessing Agreement Between Two Methods of
  Clinical Measurement}.
\newblock \emph{The Lancet}, 1\penalty0 (8476):\penalty0 307--310, 1986.

\bibitem[DeLong et~al.(1988)DeLong, DeLong, and Clarke-Pearson]{DeLong:1988gd}
Elizabeth~R DeLong, David~M DeLong, and Daniel~L Clarke-Pearson.
\newblock {Comparing the Areas under Two or More Correlated Receiver Operating
  Characteristic Curves: A Nonparametric Approach}.
\newblock \emph{Biometrics}, 44\penalty0 (3):\penalty0 837, 1988.

\bibitem[Abadi et~al.(2016)Abadi, Agarwal, Barham, Brevdo, Chen, Citro,
  Corrado, Davis, Dean, Devin, Ghemawat, Goodfellow, Harp, Irving, Isard, Jia,
  J{\'{o}}zefowicz, Kaiser, Kudlur, Levenberg, Man{\'{e}}, Monga, Moore,
  Murray, Olah, Schuster, Shlens, Steiner, Sutskever, Talwar, Tucker,
  Vanhoucke, Vasudevan, Vi{\'{e}}gas, Vinyals, Warden, Wattenberg, Wicke, Yu,
  and Zheng]{Abadi:CPnTLmNd}
Mart{\'{\i}}n Abadi, Ashish Agarwal, Paul Barham, Eugene Brevdo, Zhifeng Chen,
  Craig Citro, Gregory~S. Corrado, Andy Davis, Jeffrey Dean, Matthieu Devin,
  Sanjay Ghemawat, Ian~J. Goodfellow, Andrew Harp, Geoffrey Irving, Michael
  Isard, Yangqing Jia, Rafal J{\'{o}}zefowicz, Lukasz Kaiser, Manjunath Kudlur,
  Josh Levenberg, Dan Man{\'{e}}, Rajat Monga, Sherry Moore, Derek~Gordon
  Murray, Chris Olah, Mike Schuster, Jonathon Shlens, Benoit Steiner, Ilya
  Sutskever, Kunal Talwar, Paul~A. Tucker, Vincent Vanhoucke, Vijay Vasudevan,
  Fernanda~B. Vi{\'{e}}gas, Oriol Vinyals, Pete Warden, Martin Wattenberg,
  Martin Wicke, Yuan Yu, and Xiaoqiang Zheng.
\newblock Tensorflow: Large-scale machine learning on heterogeneous distributed
  systems.
\newblock \emph{arXiv:1603.04467 [cs.DC]}, abs/1603.04467, 2016.

\bibitem[van~der Walt et~al.(2014)van~der Walt, {S}ch\"onberger,
  {Nunez-Iglesias}, {B}oulogne, {W}arner, {Y}ager, {G}ouillart, {Y}u, and the
  scikit-image contributors]{vanderWalt:2014eo}
{S}t\'efan van~der Walt, {J}ohannes~{L}. {S}ch\"onberger, {J}uan
  {Nunez-Iglesias}, {F}ran\c{c}ois {B}oulogne, {J}oshua~{D}. {W}arner, {N}eil
  {Y}ager, {E}mmanuelle {G}ouillart, {T}ony {Y}u, and the scikit-image
  contributors.
\newblock scikit-image: image processing in {P}ython.
\newblock \emph{PeerJ}, 2:\penalty0 e453, 6 2014.
\newblock ISSN 2167-8359.
\newblock \doi{10.7717/peerj.453}.

\bibitem[Hancock et~al.(2009)Hancock, Deal, Mirvis, Okin, Kligfield, and
  Gettes]{Hancock:2009hu}
E~W Hancock, B~J Deal, D~M Mirvis, P~Okin, P~Kligfield, and L~S Gettes.
\newblock {AHA/ACCF/HRS Recommendations for the Standardization and
  Interpretation of the Electrocardiogram: Part V: Electrocardiogram Changes
  Associated With Cardiac Chamber Hypertrophy: A Scientific Statement From the
  American Heart Association Electrocardiography and Arrhythmias Committee,
  Council on Clinical Cardiology; the American College of Cardiology
  Foundation; and the Heart Rhythm Society: Endorsed by the International
  Society for Computerized Electrocardiology}.
\newblock \emph{Circulation}, 119\penalty0 (10):\penalty0 e251--e261, March
  2009.

\bibitem[Rajpurkar et~al.(2017{\natexlab{b}})Rajpurkar, Irvin, Zhu, Yang,
  arXiv, and {2017}]{Rajpurkar:tu}
P~Rajpurkar, J~Irvin, K~Zhu, B~Yang, H~Mehta arXiv~preprint arXiv, and {2017}.
\newblock {CheXNet: Radiologist-Level Pneumonia Detection on Chest X-Rays with
  Deep Learning}.
\newblock \emph{arXiv:1712.06957[physics.med-ph]}, December 2017{\natexlab{b}}.

\bibitem[Geras et~al.(2017)Geras, Wolfson, Shen, Kim, Moy, and
  Cho]{Geras:2017um}
Krzysztof~J Geras, Stacey Wolfson, Yiqiu Shen, S~Gene Kim, Linda Moy, and
  KyungHyun Cho.
\newblock {High-Resolution Breast Cancer Screening with Multi-View Deep
  Convolutional Neural Networks}.
\newblock \emph{arXiv: 1703.07047[cs.CV]}, November 2017.

\bibitem[Rajpurkar et~al.(2017{\natexlab{c}})Rajpurkar, Irvin, Bagul, Ding,
  Duan, Mehta, Yang, Zhu, Laird, Ball, Langlotz, Shpanskaya, Lungren, and
  Ng]{Rajpurkar:2017tg}
Pranav Rajpurkar, Jeremy Irvin, Aarti Bagul, Daisy Ding, Tony Duan, Hershel
  Mehta, Brandon Yang, Kaylie Zhu, Dillon Laird, Robyn~L Ball, Curtis Langlotz,
  Katie Shpanskaya, Matthew~P Lungren, and Andrew~Y Ng.
\newblock {MURA: Large Dataset for Abnormality Detection in Musculoskeletal
  Radiographs}.
\newblock \emph{arXiv:1712.06957[physics.med-ph]}, December 2017{\natexlab{c}}.

\bibitem[Gao et~al.(2017)Gao, Li, Loomes, and Lianyi]{Gao:2017fk}
Xiaohong Gao, Wei Li, Martin Loomes, and Wang Lianyi.
\newblock {A fused deep learning architecture for viewpoint classification of
  echocardiography}.
\newblock \emph{Information Fusion}, 36\penalty0 (C):\penalty0 103--113, July
  2017.

\bibitem[Madani et~al.(2018)Madani, Arnaout, Mofrad, and
  Arnaout]{Madani:2018hm}
Ali Madani, Ramy Arnaout, Mohammad Mofrad, and Rima Arnaout.
\newblock {Fast and accurate view classification of echocardiograms using deep
  learning}.
\newblock \emph{npj Digital Medicine}, 1\penalty0 (1):\penalty0 6, March 2018.

\bibitem[Lyon et~al.(2018{\natexlab{a}})Lyon, Minchol{\'e}, Mart{\'\i}nez,
  Laguna, and Rodriguez]{Lyon:2018cv}
Aurore Lyon, Ana Minchol{\'e}, Juan~Pablo Mart{\'\i}nez, Pablo Laguna, and
  Blanca Rodriguez.
\newblock {Computational techniques for ECG analysis and interpretation in
  light of their contribution to medical advances}.
\newblock \emph{Journal of The Royal Society Interface}, 15\penalty0
  (138):\penalty0 20170821--18, January 2018{\natexlab{a}}.

\bibitem[Hong et~al.(2017)Hong, Wu, Zhou, Wang, Shang, Li, and
  Xie]{Hong:2017dc}
Shenda Hong, Meng Wu, Yuxi Zhou, Qingyun Wang, Junyuan Shang, Hongyan Li, and
  Junqing Xie.
\newblock {ENCASE: an ENsemble ClASsifiEr for ECG Classification Using Expert
  Features and Deep Neural Networks}.
\newblock In \emph{2017 Computing in Cardiology Conference}. Computing in
  Cardiology, September 2017.

\bibitem[Xiong et~al.(2017)Xiong, Stiles, and Zhao]{Xiong:2017bb}
Zhaohan Xiong, Martin Stiles, and Jichao Zhao.
\newblock {Robust ECG Signal Classification for the Detection of Atrial
  Fibrillation Using Novel Neural Networks}.
\newblock In \emph{2017 Computing in Cardiology Conference}, pages 1--4.
  Computing in Cardiology, September 2017.

\bibitem[Acharya et~al.(2018)Acharya, Fujita, Oh, Hagiwara, Tan, Adam, and
  Tan]{Acharya:2018jq}
U~Rajendra Acharya, Hamido Fujita, Shu~Lih Oh, Yuki Hagiwara, Jen~Hong Tan,
  Muhammad Adam, and Ru~San Tan.
\newblock {Deep convolutional neural network for the automated diagnosis of
  congestive heart failure using ECG signals}.
\newblock \emph{Applied Intelligence}, 62\penalty0 (16):\penalty0 1495--18,
  April 2018.

\bibitem[Nahar et~al.(2013)Nahar, Imam, Tickle, and Chen]{Nahar:2013fh}
Jesmin Nahar, Tasadduq Imam, Kevin~S Tickle, and Yi-Ping~Phoebe Chen.
\newblock {Computational intelligence for heart disease diagnosis: A medical
  knowledge driven approach}.
\newblock \emph{Expert Systems With Applications}, 40\penalty0 (1):\penalty0
  96--104, January 2013.

\bibitem[Cheng et~al.(2012)Cheng, Zhu, Tian, Zhao, Cui, and Fang]{Cheng:2012dt}
Zhongwei Cheng, Kongbo Zhu, Zhuang Tian, Dachun Zhao, Quancai Cui, and Quan
  Fang.
\newblock {The Findings of Electrocardiography in Patients with Cardiac
  Amyloidosis}.
\newblock \emph{Annals of Noninvasive Electrocardiology}, 18\penalty0
  (2):\penalty0 157--162, November 2012.

\bibitem[Dumont(2006)]{Dumont:2006ee}
C~A Dumont.
\newblock {Interpretation of electrocardiographic abnormalities in hypertrophic
  cardiomyopathy with cardiac magnetic resonance}.
\newblock \emph{European Heart Journal}, 27\penalty0 (14):\penalty0 1725--1731,
  January 2006.

\bibitem[Rahman et~al.(2015)Rahman, Tereshchenko, Kongkatong, Abraham, Abraham,
  and Shatkay]{Rahman:2015jw}
Quazi~Abidur Rahman, Larisa~G Tereshchenko, Matthew Kongkatong, Theodore
  Abraham, M~Roselle Abraham, and Hagit Shatkay.
\newblock {Utilizing ECG-Based Heartbeat Classification for Hypertrophic
  Cardiomyopathy Identification.}
\newblock \emph{IEEE transactions on nanobioscience}, 14\penalty0 (5):\penalty0
  505--512, July 2015.

\bibitem[Lyon et~al.(2018{\natexlab{b}})Lyon, Ariga, Minchol{\'e}, Mahmod,
  Ormondroyd, Laguna, De~Freitas, Neubauer, Watkins, and
  Rodriguez]{Lyon:2018bk}
Aurore Lyon, Rina Ariga, Ana Minchol{\'e}, Masliza Mahmod, Elizabeth
  Ormondroyd, Pablo Laguna, Nando De~Freitas, Stefan Neubauer, Hugh Watkins,
  and Blanca Rodriguez.
\newblock {Distinct ECG Phenotypes Identified in Hypertrophic Cardiomyopathy
  Using Machine Learning Associate With Arrhythmic Risk Markers}.
\newblock \emph{Frontiers in Physiology}, 9:\penalty0 1369--13, March
  2018{\natexlab{b}}.

\bibitem[Fikrle et~al.(2013)Fikrle, Pale{\v c}ek, Kuchynka, N{\v e}me{\v c}ek,
  Bauerov{\'a}, Straub, and Ry{\v s}av{\'a}]{Fikrle:2013de}
Michal Fikrle, Tom{\'a}{\v s} Pale{\v c}ek, Petr Kuchynka, Eduard N{\v e}me{\v
  c}ek, Lenka Bauerov{\'a}, Jan Straub, and Romana Ry{\v s}av{\'a}.
\newblock {Cardiac amyloidosis: A comprehensive review}.
\newblock \emph{Cor et Vasa}, 55\penalty0 (1):\penalty0 e60--e75, February
  2013.

\bibitem[Kopec et~al.(2012)Kopec, Tyrka, Miszalski-Jamka, Sobien, Walig~oacute
  ra, Br~oacute zda, and Podolec]{Kopec:2012bp}
Grzegorz Kopec, Anna Tyrka, Tomasz Miszalski-Jamka, Maciej Sobien, Marcin
  Walig~oacute ra, Mateusz Br~oacute zda, and Piotr Podolec.
\newblock {Electrocardiogram for the Diagnosis of Right Ventricular Hypertrophy
  and Dilation in Idiopathic Pulmonary Arterial Hypertension}.
\newblock \emph{Circulation journal: official journal of the Japanese
  Circulation Society}, 76\penalty0 (7):\penalty0 1744--1749, 2012.

\bibitem[Peighambari et~al.(2014)Peighambari, Alizadehasl, and
  Totonchi]{Peighambari:2014ib}
Mohammad~Mehdi Peighambari, Azin Alizadehasl, and Ziae Totonchi.
\newblock {Electrocardiographic changes in mitral valve prolapse syndrome.}
\newblock \emph{Journal of cardiovascular and thoracic research}, 6\penalty0
  (1):\penalty0 21--23, 2014.

\bibitem[Sengupta et~al.(2018)Sengupta, Kulkarni, and Narula]{Sengupta:2018hl}
Partho~P Sengupta, Hemant Kulkarni, and Jagat Narula.
\newblock {Prediction of Abnormal Myocardial Relaxation From Signal Processed
  Surface ECG}.
\newblock \emph{Journal of the American College of Cardiology}, 71\penalty0
  (15):\penalty0 1650--1660, April 2018.

\bibitem[Levy et~al.(1994)Levy, Salomon, D'Agostino, Belanger, and
  Kannel]{Levy:1994ue}
D~Levy, M~Salomon, R~B D'Agostino, A~J Belanger, and W~B Kannel.
\newblock {Prognostic implications of baseline electrocardiographic features
  and their serial changes in subjects with left ventricular hypertrophy.}
\newblock \emph{Circulation}, 90\penalty0 (4):\penalty0 1786--1793, October
  1994.

\bibitem[Tonelli et~al.(2013)Tonelli, Baumgartner, Alkukhun, Minai, and
  Dweik]{Tonelli:2013ii}
Adriano~R Tonelli, Manfred Baumgartner, Laith Alkukhun, Omar~A Minai, and
  Raed~A Dweik.
\newblock {Electrocardiography at Diagnosis and Close to the Time of Death in
  Pulmonary Arterial Hypertension}.
\newblock \emph{Annals of Noninvasive Electrocardiology}, 19\penalty0
  (3):\penalty0 258--265, December 2013.

\bibitem[Henkens et~al.(2008)Henkens, Gan, van Wolferen, Hew, Boonstra, Twisk,
  Kamp, van~der Wall, Schalij, Noordegraaf, and Vliegen]{Henkens:2008ky}
Ivo~R Henkens, C~Tji-Joong Gan, Serge~A van Wolferen, Miki Hew, Anco Boonstra,
  Jos W~R Twisk, Otto Kamp, Ernst~E van~der Wall, Martin~J Schalij, Anton~Vonk
  Noordegraaf, and Hubert~W Vliegen.
\newblock {ECG Monitoring of Treatment Response in Pulmonary Arterial
  Hypertension Patients}.
\newblock \emph{Chest}, 134\penalty0 (6):\penalty0 1250--1257, December 2008.

\bibitem[Walig{\'o}ra et~al.(2018)Walig{\'o}ra, Tyrka, Podolec, and
  Kopec]{Waligora:2018ig}
Marcin Walig{\'o}ra, Anna Tyrka, Piotr Podolec, and Grzegorz Kopec.
\newblock {ECG Markers of Hemodynamic Improvement in Patients with Pulmonary
  Hypertension}.
\newblock \emph{BioMed Research International}, 2018\penalty0 (1):\penalty0
  1--10, 2018.

\bibitem[Laguna et~al.(1994)Laguna, Jan{\'e}, and Caminal]{Laguna:1994up}
P~Laguna, R~Jan{\'e}, and P~Caminal.
\newblock {Automatic detection of wave boundaries in multilead ECG signals:
  validation with the CSE database.}
\newblock \emph{Computers and biomedical research, an international journal},
  27\penalty0 (1):\penalty0 45--60, February 1994.

\bibitem[Brohet et~al.(1984)Brohet, Robert, Derwael, Fesler, Stijns, Vliers,
  and Braasseur]{Brohet:1984ch}
C~R Brohet, A~Robert, C~Derwael, R~Fesler, M~Stijns, A~Vliers, and L~A
  Braasseur.
\newblock {Computer interpretation of pediatric orthogonal electrocardiograms:
  statistical and deterministic classification methods}.
\newblock \emph{Circulation}, 70\penalty0 (2):\penalty0 255--262, 1984.

\bibitem[Kors and van Bemmel(1990)]{Kors:1990wn}
J~A Kors and J~H van Bemmel.
\newblock {Classification methods for computerized interpretation of the
  electrocardiogram.}
\newblock \emph{Methods of information in medicine}, 29\penalty0 (4):\penalty0
  330--336, September 1990.

\bibitem[Kingma and Ba(2014)]{Kingma:2014us}
Diederik~P Kingma and Jimmy Ba.
\newblock {Adam: A Method for Stochastic Optimization.}
\newblock \emph{arXiv:1412.6980 [cs.LG]}, December 2014.

\end{thebibliography}

\onecolumn
\newpage

\captionsetup*{format=largeformat}

\beginsupplement

\section{Convolutional Neural Networks for Segmentation} \label{note:Note1} 
Our implementation for segmentation is based on the U-net architecture \cite{Ronneberger:2015vw}, illustrated in Figure \ref{fig:figure2}, but novel in its adaptation to 1-D feature vector inputs. The architecture consists of "Convolution Blocks N" ("CB N" for short), which are repeated groups of 3 consecutive convolution layers that apply 1xN filters with the "same" padding. The network takes an input of 2000x12 images (12 leads each sampled to 2000 pixels) that is fed into two sets of contracting and expanding paths. The first set of contracting and expanding paths consists of 7 convolution blocks that applies convolutions with filter sizes starting from 1x19 to 1x15 to 1x11. The contracting path is implemented by applying 2x2 max pool after the first 5 convolution blocks and the expanding path is implemented by applying an 8x8 deconvolution layer and a 2x2 deconvolution layer after the next two convolution blocks respectively. The intuition for this set of convolutions is to independently detect features between the 12 leads by having convolution filters that only slide along one lead at a time. A 12x1 convolution with valid padding is then applied to collapse the 12 lead vectors into 1 feature vector. This feeds into the second set of contracting and expanding path, which consists of a convolution block with 1x11 filters followed by a 2x2 max pool and a convolution block with 1x11 filters followed by a 2x2 deconvolution layer. One more convolution block with 1x9 filters is applied before we create a 6 class segmentation output. All deconvolution outputs in the expanding phase are concatenated with feature maps from convolution block outputs in the contracting phase with the same image dimensions. 

For stochastic optimization, we used the ADAM optimizer \cite{Kingma:2014us} with an initial learning rate of 1x10$^{-5}$ and mini-batch size of 5. For regularization, we applied a weight decay of 1x10$^{-7}$ on all network weights and dropout with probability 0.5 on the final convolution block. We ran our tests for 500 epochs, which takes ~4 hours on a Nvidia GTX 1080. Accuracy was assessed by 5-fold cross-validation.

\newpage

\begin{table*}[] 
   \small 
   \centering 
   \begin{tabular}{| l | l | c | c | } %
   \hline
   \multirow{1}{*}    && \textbf{All Patients} 	& \textbf{Patients Used for Model} \\
   \hline
      \multirow{1}{*}{\textbf{Characteristics}}& Age (years $\pm$ sd) & 58 $\pm$ 17 & 58 $\pm$ 17\\
                                     & Sex (\% Female) & 48 & 49\\
    & Left Ventricular Mass Index (g/m$^2$) & 85 $\pm$ 29 & 84 $\pm$ 26 \\
   \hline
      \multirow{1}{*}{\textbf{Race (\%)}}
   & Unknown/Other   					   & 54 & 55\\
   & Non-Hispanic White                  & 32 & 32 \\
   & Black 								& 6 & 6 \\
   & Asian and Pacific Islander 		& 6 & 6  \\
   & Hispanic 							& 1 & 1  \\
   & Native American 					& <1 & <1 \\
   \hline
   \multirow{1}{*}{\textbf{Year}} 		 & 2017 & 3550 (35) & 3064 (35)\\
                                    	 & 2016 & 3241 (32) & 2732 (32)\\
                                         & 2015 & 2356 (23) & 2033 (24) \\
                                         & 2014 & 935  (9) &  802 (9) \\
                                         & \textbf{Total}  & \textbf{10082} & \textbf{8631} \\
   \hline
   \end{tabular}
   \caption[]{\textbf{Characteristics of Studies Used to Train Left Ventricular Mass Estimation Model}} \label{tab:tableS1}
\end{table*}

\begin{table*}[] 
   \small 
   \centering 
   \begin{tabular}{| l | l | c | c | } %
   \hline
   \multirow{1}{*}    && \textbf{All Patients} 	& \textbf{Patients Used for Model} \\
   \hline
      \multirow{1}{*}{\textbf{Characteristics}}& Age (years $\pm$ sd) & 59 $\pm$ 17 & 58 $\pm$ 17\\
                                     & Sex (\% Female) & 48 & 48\\
   & Left Atrial Volume Index (mL/m$^2$) & 30 $\pm$ 13 & 29 $\pm$ 12 \\ 
   \hline
      \multirow{1}{*}{\textbf{Race (\%)}}
   & Unknown/Other                      & 53 &  53 \\
   & Non-Hispanic White                 & 33 & 33 \\
   & Black 								& 7 & 6 \\
   & Asian and Pacific Islander 		& 6 & 6 \\
   & Hispanic 							& 1 & 1 \\
   & Native American 					& <1 & <1 \\
   \hline
   \multirow{1}{*}{\textbf{Year}} 		 & 2017 & 2888 (31) & 2223 (32) \\
                                    	 & 2016 & 2611 (35) & 2431 (34) \\
                                         & 2015 & 2084 (25) & 1797 (25) \\
                                         & 2014 & 706 (9) & 602 (9) \\
                                         & \textbf{Total} & \textbf{8289} & \textbf{7053} \\
   \hline
   \end{tabular}
   \caption[]{\textbf{Characteristics of Studies Used to Train Left Atrial Volume Estimation Model}} \label{tab:tableS2}
\end{table*}

\begin{table*}[] 
   \small 
   \centering 
   \begin{tabular}{| l | l | c | c | } %
   \hline
      \multirow{1}{*}    && \textbf{All Patients} 	& \textbf{Patients Used for Model} \\
      \hline
      \multirow{1}{*}{\textbf{Characteristics}}& Age (years $\pm$ sd) & 59$\pm$ 17 & 58 $\pm$ 17\\
                                     & Sex (\% Female) & 48 & 49 \\
    & Mitral valve medial e' (cm/s) & 0.074 $\pm$ 0.026 & 0.074 $\pm$ 0.025 \\
   \hline
   
      \multirow{1}{*}{\textbf{Race (\%)}}
   & Unknown/Other    					& 53&  53 \\
   & Non-Hispanic White                 & 34& 34 \\
   & Black 								& 7& 7 \\
   & Asian and Pacific Islander 		& 6& 6 \\
   & Hispanic 							& 1& 1 \\
   & Native American 					& <1&  <1 \\
   \hline
   \multirow{1}{*}{\textbf{Year}} 		 & 2017 & 963 (23) & 847 (23) \\
                                    	 & 2016 & 1674 (40) & 1420 (39) \\
                                         & 2015 & 1121 (27) & 971 (27) \\
                                         & 2014 & 447 (11) & 391 (11) \\
                                         & \textbf{Total} & \textbf{4205} & \textbf{3629} \\
   \hline
   \end{tabular}
   \caption[]{\textbf{Characteristics of Studies Used to Train Mitral Valve Annular e' Estimation Model}} \label{tab:tableS3}
\end{table*}


\begin{table*}[] 
   \small 
   \centering 
   \begin{tabular}{| l | c | c | c | } %
   \hline
   \multirow{1}{*}    && \textbf{Cases} 	& \textbf{Controls} \\
   \hline
      \multirow{1}{*}{\textbf{Patients}}
   & N (studies) & 525 & 1764 \\
   & Age (years $\pm$ sd) & 52 $\pm$ 11 & 52 $\pm$ 11 \\
   & Sex (\% Female) & 64 &  65  \\
   \hline
      \multirow{1}{*}{\textbf{Race (\%)}}
   & Unknown/Other     & 49 & 47   \\
   & Non-Hispanic White                    & 44 & 45  \\
   & Black 	& 4  & 5 \\
   & Asian and Pacific Islander    &  1 & 1  \\
   & Hispanic & 2  & 3   \\
   \hline

   \multirow{1}{*}{\textbf{Year - Number of Studies (\%)}} 
   										 & 2017 & 149 (28) & 475 (27) \\
   										 & 2016 & 194 (37) & 639 (36) \\
   										 & 2015 & 68 (13) & 234 (13) \\
   										 & 2014 & 52 (10) & 192 (11) \\
   										 & 2011 & 31 (6) & 114 (6) \\
                     & 2010  & 31 (6) & 110 (6) \\
   \hline

   \end{tabular}
   \caption[]{\textbf{Characteristics of Studies Used to Train PAH Classification Model}}\label{tab:tableS4}
\end{table*}


\begin{table*}[] 
   \small 
   \centering 
   \begin{tabular}{| l | c | c | c | } %
   \hline
   \multirow{1}{*}    && \textbf{Cases} 	& \textbf{Controls} \\
   \hline
      \multirow{1}{*}{\textbf{Patients}}
   & N (studies) & 706 & 2577 \\
   & Age (years $\pm$ sd) & 57 $\pm$ 15 & 58 $\pm$ 15 \\
   & Sex (\% Female) & 47 & 46  \\
   & Genotype Positive (\%)  & 18 & 0 \\
   \hline
      \multirow{1}{*}{\textbf{Race (\%)}}
   & Unknown/Other     & 44 & 41   \\
   & Non-Hispanic White                    & 44 & 47  \\
   & Black 	& 6  & 5 \\
   & Asian and Pacific Islander    &  3 & 3  \\
   & Hispanic & 3  & 2   \\
   \hline

   \multirow{1}{*}{\textbf{Year - Number of Studies (\%)}} 
   										 & 2017 & 151 (21) & 557 (22) \\
   										 & 2016 & 160 (23) & 557 (22) \\
   										 & 2015 & 112 (16) & 412 (16) \\
   										 & 2014 & 133 (19) & 475 (18) \\
   										 & 2011 & 88 (12) & 352 (13) \\
                               & 2010  & 62 (9) & 224 (9) \\
   \hline

   \end{tabular}
   \caption[]{\textbf{Characteristics of Studies Used to Train HCM Classification Model}}\label{tab:tableS5}
\end{table*}


\begin{table*}[] 
   \small 
   \centering 
   \begin{tabular}{| l | c | c | c | } %
   \hline
   \multirow{1}{*}    && \textbf{Cases} 	& \textbf{Controls} \\
   \hline
      \multirow{1}{*}{\textbf{Patients}}
   & N (studies) & 180 & 636 \\
   & Age (years) & 67 $\pm$ 11 & 67 $\pm$ 11 \\
   & Sex (\% Female) & 14 & 14  \\
   \hline
      \multirow{1}{*}{\textbf{Race (\%)}}
   & Unknown/Other     & 26 & 26   \\
   & Non-Hispanic White                    & 49 & 50  \\
   & Black 	& 20  & 18 \\
   & Asian and Pacific Islander    &  4 & 4  \\
   & Hispanic & <1  & <1   \\
   \hline

   \multirow{1}{*}{\textbf{Year - Number of Studies (\%)}} 
   										 & 2017 & 32 (18) & 122 (19) \\
   										 & 2016 & 32 (18) & 122 (19) \\
   										 & 2015 & 33 (18) & 124 (19) \\
   										 & 2014 & 19 (11) & 74 (12) \\
   										 & 2011 & 34 (19) & 108 (17) \\
                               & 2010  & 30 (13) & 86 (14) \\
   \hline

   \end{tabular}
   \caption[]{\textbf{Characteristics of Studies Used to Train Amyloid Classification Model}} \label{tab:tableS6}
\end{table*}


\begin{table*}[] 
   \small 
   \centering 
   \begin{tabular}{| l | c | c | c | } %
   \hline
   \multirow{1}{*}    && \textbf{Cases} 	& \textbf{Controls} \\
   \hline
      \multirow{1}{*}{\textbf{Patients}}
   & N (studies) & 1576 & 7493 \\
   & Age (years $\pm$ sd) & 61 $\pm$ 15 & 61 $\pm$ 15 \\
   & Sex (\% Female) & 49 & 49  \\
   \hline
      \multirow{1}{*}{\textbf{Race (\%)}}
   & Unknown/Other     & 36 & 36   \\
   & Non-Hispanic White                    & 56 & 57  \\
   & Black 	& 2  & 2 \\
   & Asian and Pacific Islander    &  5 & 4  \\
   & Hispanic & <1  & <1   \\
   \hline

   \multirow{1}{*}{\textbf{Year - Number of Studies (\%)}} 
   										 & 2017 & 307 (19) & 979 (20) \\
   										 & 2016 & 415 (26) & 1302 (27) \\
   										 & 2015 & 359 (23) & 1050 (22) \\
   										 & 2014 & 303 (19) & 907 (19) \\
   										 & 2011 & 80 (5) & 269 (6) \\
                               & 2010  & 112 (7) & 357 (7) \\
   \hline

   \end{tabular}
   \caption[]{\textbf{Characteristics of Studies Used to Train MVP Classification Model}} \label{tab:tableS7}
\end{table*}


\begin{table*}[] 
   \small 
   \centering 
   \begin{tabular}{| l | c | c | c | c | } %
   \hline
   \multirow{1}{*}    && \textbf{Cases} 	& \textbf{Controls} & \textbf{pvalue}\\
   \hline
      \multirow{1}{*}{\textbf{Patients}}
   & N (studies) & 853 & 4272 & \\
   & Age (years $\pm$ sd) & 62 $\pm$ 16 & 56 $\pm$ 17 & 9x10$^{-22}$\\
   & Sex (\% Female) & 49 & 49  & \\
   & BMI (kg/m$^2$, IQR) & 29 (23-31) & 27 (22-30) & 1x10$^{-5}$ \\
   & LV Mass Index (g/m$^2$, IQR) & 139 (124-151) & 65 (58-73) & 0 \\
   \hline
      \multirow{1}{*}{\textbf{Race (\%)}}
   & Unknown/Other     & 52 & 57  & \\
   & Non-Hispanic White                    & 29 & 32  & \\
   & Black 	& 13 & 4 & 3x10$^{-18}$\\
   & Asian and Pacific Islander    &  6 & 6 &  \\
   & Hispanic & 1  & 1  &  \\
   \hline

   \multirow{1}{*}{\textbf{Year - Number of Studies (\%)}} 
   										 & 2017 & 304 (36) & 1504 (35) & \\
   										 & 2016 & 272 (32) & 1323 (31) & 0.1\\
   										 & 2015 & 170 (20) & 1077 (25) & \\
   										 & 2014 & 109 (13) & 368 (9) & \\
   \hline

   \end{tabular}
   \caption[]{\textbf{Characteristics of Studies Used to Train Left Ventricular Hypertrophy Classification Model}} 
   \label{tab:tableS8}
\end{table*}


\begin{table*}[] 
   \small 
   \centering 
   \begin{tabular}{| l | c | c | c | c | } %
   \hline
   \multirow{1}{*}    && \textbf{Cases} 	& \textbf{Controls} & \textbf{pvalue}\\
   \hline
      \multirow{1}{*}{\textbf{Patients}}
   & N (studies) & 362 & 1823 & \\
   & Age (years $\pm$ sd) & 69 $\pm$ 15 & 51 $\pm$ 16 & 5x10$^{-74}$\\
   & Sex (\% Female) & 49 & 49  & \\
   & BMI (kg/m$^2$, IQR) & 27 (22-29) & 27 (22-30) & 0.49 \\
   & MV annulus e' (cm/s, IQR) & 0.038 (0.034-0.042) & 0.093 (0.079-0.104) & 0\\
   \hline
      \multirow{1}{*}{\textbf{Race (\%)}}
   & Unknown/Other     & 52 & 54  & \\
   & Non-Hispanic White                    & 32 & 35  & \\
   & Black 	& 9  & 6 & 9x10$^{-38}$\\
   & Asian and Pacific Islander    &  7 & 4 &  \\
   & Hispanic & 1  & 1  &  \\
   \hline

   \multirow{1}{*}{\textbf{Year - Number of Studies (\%)}} 
   										 & 2017 & 71 (20) & 443 (24) & \\
   										 & 2016 & 138 (38) & 717 (39) & 7x10$^{-7}$\\
   										 & 2015 & 104 (29) & 471 (26) & \\
   										 & 2014 & 49 (14) & 192 (11) & \\
   \hline

   \end{tabular}
   \caption[]{\textbf{Characteristics of Studies Used to Train Diastolic Dysfunction Classification Model}} \label{tab:tableS9}
\end{table*}


\begin{table*}[] 
   \small 
   \centering 
   \begin{tabular}{| l | c | c | c | c | } %
   \hline
   \multirow{1}{*}    && \textbf{Cases} 	& \textbf{Controls} & \textbf{pvalue}\\
   \hline
      \multirow{1}{*}{\textbf{Patients}}
   & N (studies) & 699 & 3490 & \\
   & Age (years $\pm$ sd) & 66 $\pm$ 16 & 55 $\pm$ 17 & 6x10$^{-53}$\\
   & Sex (\% Female) & 48 & 48  & \\
   & BMI (kg/m$^2$, IQR) & 28 (23-30) & 28 (23-30) & 0.41 \\
   & Left atrial volume index (mL/m$^2$, IQR) & 55 (48-59) & 20 (18-24) & 0\\
   \hline
      \multirow{1}{*}{\textbf{Race (\%)}}
   & Unknown/Other     & 49 & 55  & \\
   & Non-Hispanic White                    & 35 & 32  & \\
   & Black 	& 10  & 6 & 0.03\\
   & Asian and Pacific Islander    &  5 & 6 &  \\
   & Hispanic & 1  & 1  &  \\
   \hline

   \multirow{1}{*}{\textbf{Year - Number of Studies (\%)}} 
   										 & 2017 & 263 (38) & 1206 (986) & \\
   										 & 2016 & 241 (34) & 991 (35) & 0.1\\
   										 & 2015 & 150 (21) & 986 (28) & \\
   										 & 2014 & 45 (6) & 307 (9) & \\
   \hline

   \end{tabular}
   \caption[]{\textbf{Characteristics of Studies Used to Train Left Atrial Enlargement Classification Model}} \label{tab:tableS10}
\end{table*}


\begin{table*}[] 
   \small 
   \centering 
   \begin{tabular}{| l | l | c |} %
   \hline
   \multirow{1}{*}{\textbf{Metric}} 				&  & \textbf{Variable Score} \\
   \hline
   \multirow{1}{*}{\textbf{LV Mass Index}} 				
       & QRS duration                          & 4.0 \\
       & P wave duration                        & 3.3 \\
       & QT duration                           & 1.7 \\
       & QRS V3 seg 8-12                   & 1.5  \\
       & ST-T v1 12-16                          & 1.3 \\
       & ST-T v1 8-12                          & 1.2 \\
       & QRS V3 seg 12-16                  & 0.84 \\
       & QRS aVL seg 12-16                  & 0.84  \\
       & QRS V6 seg 8-12                  & 0.75  \\
   \hline
   \multirow{1}{*}{\textbf{LA Volume Index}} 				
       & QT duration                           & 4.6 \\
       & P wave duration                        & 4.5 \\
       & QRS duration                          & 1.4 \\
       & PR duration                          & 1.3 \\
       & QRS V6 seg 12-16                 & 0.97  \\
       & ST-T v1 8-12                          & 0.88 \\
       & QRS V3 seg 8-12                 & 0.76  \\
   \hline
   \multirow{1}{*}{\textbf{MV medial e'}} 				
       & PR duration                          & 3.1 \\
       & QT duration                           & 2.9 \\
       & P wave duration                           & 2.4 \\
       & ST-T V1 8-12                          & 1.8 \\
       & Heart rate                         & 1.2 \\
       & QRS aVF seg 8-12                 & 1.1  \\
       & QRS V6 seg 0-4                 & 0.91  \\
   \hline
   \end{tabular}
   \caption[]{\textbf{Variable Importance for GBM Models for Cardiac Structure and Function}}\label{tab:tableS11}
\end{table*}


\begin{table*}[] 
   \small 
   \centering 
   \begin{tabular}{| l | l | c |} %
   \hline
   \multirow{1}{*}{\textbf{Metric}} 				&  & \textbf{Variable Score} \\
   \hline
   \multirow{1}{*}{\textbf{PAH}} 				
       & QRS V1 seg 8-12                  & 4.5  \\
       & QRS V1 seg 12-16                 & 1.6  \\
       & QRS I seg 12-16                  & 1.4  \\
       & P-PR V3 seg 4-8                  & 0.93 \\
       & P-PR aVR seg 4-8                 & 0.93 \\
       & QRS I seg 8-12                   & 0.91 \\
       & QRS duration                     & 0.87 \\
       & QRS aVR seg 8-12                  & 0.85 \\
   \hline
   \multirow{1}{*}{\textbf{HCM}} 				
       & ST-T V1 seg 12-16                         & 3.8 \\
       & P wave duration                        & 3.5 \\
       & QT duration                           & 2.7 \\
       & PR duration                          & 2.4 \\
       & QRS aVR seg 12-16                 & 1.3  \\
       & ST-T V1 seg 8-12                 & 1.2  \\
       & ST-T V5 seg 8-12                 & 0.83  \\
       & Heart rate                          & 0.76\\
   \hline
   \multirow{1}{*}{\textbf{Amyloid}} 				
       & QRS aVR seg 4-8                   & 3.0 \\
       & QRS duration                          & 1.3 \\
       & QRS I seg 8-12                  & 1.2 \\
       & QRS I seg 4-8                   & 1.1 \\
       & QRS V1 seg 0-4                   & 1.1 \\
       & QRS aVL seg 8-12                  & 0.99 \\
       & ST-T V2 seg 12-16                & 0.91 \\
       & P wave duration                           & 0.78 \\
   \hline
   \multirow{1}{*}{\textbf{MVP}} 				
       & PR duration                          & 3.3 \\
       & QRS V2 seg 4-8                   & 1.2 \\
       & QRS V3 seg 0-4                   & 1.2 \\
       & P wave duration                           & 1.1 \\
       & QT duration                           & 0.97\\
       & P-PR V1 seg 0-4                  & 0.75 \\
   \hline
   \hline
   \end{tabular}
   \caption[]{\textbf{Variable Importance for GBM Models for Disease Detection}} \label{tab:tableS12}
\end{table*}

\end{document}